\documentclass{article}

\usepackage[final,nonatbib]{neurips_2020}

\usepackage[utf8]{inputenc} 
\usepackage[T1]{fontenc}    
\usepackage{hyperref}       
\usepackage{url}            
\usepackage{booktabs}       
\usepackage{amsfonts}       
\usepackage{nicefrac}       
\usepackage{microtype}      

\usepackage{graphicx}
\graphicspath{{images/}}

\usepackage{amsmath}  
\usepackage{amssymb}  
\usepackage{bm}       

\usepackage{tabularx}

\usepackage[tight,bf]{subfigure}

\usepackage[ruled,vlined,linesnumbered]{algorithm2e}
\usepackage{etoolbox}
\AtBeginEnvironment{algorithm}{\SetArgSty{textnormal}}  

\usepackage{appendix}

\usepackage{xcolor}
\usepackage{listings}

\lstdefinestyle{mystyle}{
    commentstyle=\color{cyan},
    keywordstyle=\color{magenta},
    stringstyle=\color{orange},
    basicstyle=\ttfamily\scriptsize,
    breakatwhitespace=false,
    breaklines=true,
    captionpos=b,
    keepspaces=true,
    showspaces=false,
    showstringspaces=false,
    showtabs=false,
    tabsize=2
}
\title{Data Generating Process to Evaluate Causal Discovery Techniques for Time Series Data}

\author{
  Andrew R.~Lawrence\thanks{Corresponding author.}\:\,\thanks{These authors contributed equally to this work.} \qquad
  Marcus Kaiser\footnotemark[\value{footnote}] \qquad
  Rui Sampaio \qquad
  Maksim Sipos \\
  causaLens, London, UK \\
  \texttt{\{andrew,marcus,rui,max\}@causalens.com} \\
}

\begin{document}

\maketitle

\begin{abstract}

  Going beyond correlations, the understanding and identification of causal relationships in observational time
  series, an important subfield of Causal Discovery, poses a major challenge.
  The lack of access to a well-defined ground truth for real-world data creates the need to rely on synthetic data
  for the evaluation of these methods.
  Existing benchmarks are limited in their scope, as they either are restricted to a ``static'' selection of data sets,
  or do not allow for a granular assessment of the methods' performance when commonly made assumptions are violated.
  We propose a flexible and simple to use framework for generating time series data,
  which is aimed at developing, evaluating, and benchmarking time series causal discovery methods.
  In particular, the framework can be used to fine tune novel methods on vast amounts of data, without ``overfitting''
  them to a benchmark, but rather so they perform well in real-world use cases.
  Using our framework, we evaluate prominent time series causal discovery methods and demonstrate a notable
  degradation in performance when their assumptions are invalidated and their sensitivity to choice of
  hyperparameters.
  Finally, we propose future research directions and how our framework can support both researchers and practitioners.

\end{abstract}


\section{Introduction} \label{sec:introduction}

The aim of {\it Causal Discovery} is to identify causal relationships from purely observational data.
Special interest lies in identifying causal effects for time series data,
where individual observations arrive ordered in time.
Compared to the case of independent and identically distributed (IID) data, a robust analysis of time series data
requires one to address additional difficulties and guard against pitfalls.
These difficulties include non-stationarity, which can materialize in shifts in distribution
(e.g.,\ a shift in the mean or a higher moment, potentially from interventions).
Moreover, real-world time series tend to show various levels of autocorrelation.
These can both carry valuable long-term information, but also invalidate typical assumptions for
statistical procedures, such as the independence of samples
(cf.~the Gauss-Markov Theorem for linear regression~\cite[Chapter 3.3.2]{friedman2001elements}).

One major benefit is that the order of time can help distinguish cause from effect.
As the future cannot affect the past, the causal driver can be identified as the variable that occurred first.
This is a valid assumption for high-resolution data;
however, for lower resolutions, one must consider {\it contemporaneous} or {\it instantaneous effects},
where one variable has a causal effect on another variable observed at the same point in time.

Additional complications arise when it comes to evaluating and comparing the performance of individual methods.
Frequently, new techniques are evaluated against their own synthetic benchmarks, rather than following
one ``gold standard'' such as those which have been established in other domains within machine learning,
e.g.\ MNIST~\cite{lecun1998gradient,lecun2010mnist} and CIFAR-10/100~\cite{krizhevsky2009learning}
for image classification.
The lack of a general benchmark with known ground truth makes it difficult to compare individual methods,
especially when there is not a publicly available implementation of the new method.

For real-world problems, there is often no known ground truth causal structure and it is impossible to observe
all the variables to ensure causal sufficiency~\cite{spirtes2000causation}.
In many cases it is unclear how a causal structure can be defined.
For example, consider a system with many highly correlated time series;
in such a setup, it is often not straightforward to identify whether an observed effect stems from a single time series,
a subset, or all of them.
Moreover, real-world data often violate assumptions made for causal discovery methods (such as IID data,
or linear relationships between variables).
Therefore, it is important to test how individual methods perform when these assumptions are not satisfied.

We propose a flexible, yet easy to use synthetic data generation process based on
structural causal models~\cite{pearl2009causal},
which can be used to benchmark causal discovery methods under a wide-range of conditions.
We show the performance of prominent methods when key assumptions are invalidated and demonstrate the sensitivity
to the choice of hyperparameter values.

\section{Background} \label{sec:background}

\subsection{Brief overview of causal discovery methods} \label{sec:brief_overview}

We here give a high level overview of Causal Graph Discovery methods, which aim to identify a
{\it Directed Acyclic Graph (DAG)} from purely observational data.
A DAG consists of nodes connected by directed edges/links from parent nodes to child nodes.
Nodes in the DAG represent the variables in the data and edges indicate direct causes, i.e.,\
a variable is said to be a direct cause of another if the former is a parent of the latter in the DAG.
Classical methods for Causal Discovery for IID data
either depend on {\it Conditional Independence Tests} or are based on {\it Functional Causal Models}.
For a recent review of the topic, we refer the reader to~\cite{glymour2019review}.

Methods that depend on {\it Conditional Independence Tests} can be seen as a special case of discovery methods for
{\it Bayesian Networks}~\cite{spirtes2000constructing}.
These methods use a series of conditional independence tests to construct a graph that satisfies the
{\it Causal Markov Condition}, cf.~\cite[Chapter 3.4.1]{spirtes2000causation},~\cite{spirtes2000constructing}.
Typically, the convergence to a true DAG cannot be guaranteed and the resulting graph is only partially directed
(a {\it Completed Partially DAG (CPDAG)}, cf.~\cite{chickering2002optimal, colombo2012learning} for more details).
Two subclasses of these algorithms are {\it Constraint Based Methods} (e.g.,\ PC~\cite{spirtes2000constructing}
and (R)FCI~\cite{colombo2012learning})
and {\it Score Based Methods}, which optimize a score that results in a graph that is (as close as possible to) a DAG
(e.g.,\ Greedy Equivalence Search~(GES), see~\cite{chickering2002optimal,meek1997graphical}).
Both constraint and score based methods can be combined to obtain {\it Hybrid Methods},
such as GFCI~\cite{glymour2019review}, which combines GES with FCI.

{\it Functional Causal Models} prescribe a specific functional form to the relation between variables
(see~\autoref{sec:scm}).
A well-known example is the Linear Non-Gaussian Additive Model
(LiNGAM)~\cite{shimizu2006linear, shimizu2011directlingam}.
A more recent example within this class is NoTEARS~\cite{zheng2018dags}, which encodes a DAG-constraint
as part of a differentiable loss function.
There are non-linear extensions of NoTEARS~\cite{yu2019dag,zheng2020learning}
and similar ideas with optimization of a loss for learning DAGs have been used
in~\cite{ng2020role} and~\cite{varando2020learning}.
This class of methods returns a functional representation, from which a DAG can be obtained.
Note that the latter is typically not assumed to be ``causal'' as it may not satisfy the Causal Markov Condition.

\paragraph{Time series causal discovery}
For time series causal discovery there are two notions of a causal graph \textemdash the \textit{Full Time Graph}
(FG) and the \textit{Summary Graph} (SG); both defined in~\cite[Chapter 10.1]{peters2017elements}.
The FG is a DAG whose nodes represent the variables at each point in time, with the convention that future values cannot
be parents of present or past values.
The SG is a ``collapsed'' version of the FG, where each node represents a whole time series.
There exists an edge $X_i \rightarrow X_j$ in the SG if and only if there exists $t \le t'$ s.t.\
$X_i(t) \rightarrow X_j(t')$ in the FG.

The classical approach to causal discovery for time series is \textit{Granger Causality}~\cite{granger1969investigating}.
Intuitively, for two time series $X$ and $Y$ in a universe of observed time series $U$, we say that
``$X$ \textit{Granger causes} $Y$'' if excluding historical values of $X$ from the universe $U$ decreases the
forecasting performance of $Y$.
Non-linear versions of Granger Causality~\cite{marinazzo2008kernel} have been proposed and
Granger Causality is closely linked to (the non-linear) \textit{Transfer Entropy},
cf.~\cite{schreiber2000measuring, barnett2009granger}.
The concept has been extended to multivariate Granger Causality approaches, often combined with a
sparsity inducing Lasso penalty, cf.~\cite{arnold2007temporal,shojaie2010discovering}.

Beyond Granger Causality, there have been many recent approaches to Causal Discovery for time series, particularly
at the FG level.
PCMCI/PCMCI+~\cite{runge2018causal,Rungeeaau4996,runge2020} and the related LPCMCI~\cite{gerhardus2020high} execute a
two-step procedure.
The first step consists of estimating a set of parents for each variable, which is based on PC and FCI,
respectively.
In the second step, we test for conditional independence of any two variables conditioned on the union of their parents.
Furthermore, VAR-LiNGAM~\cite{hyvarinen2010estimation}, DYNOTEARS~\cite{pamfil2020dynotears} and
SVAR-(G)FCI~\cite{malinsky2019learning} are vector-autoregressive extensions of LiNGAM, NoTEARS and FCI / GFCI,
respectively.
Further recent references for time series causal discovery methods can be found
in~\cite{peters2013causal,entner2010causal,huang2020causal,weichwald2020causal}.

\subsection{Structural causal models} \label{sec:scm}

Next we introduce {\it Structural Causal Models} (SCM)~\cite{pearl2009causal},
also known as {\it Structural Equation Models} or {\it Functional Causal Models}.
These models assume that child nodes in a causal graph have a functional dependence on their parents.
More precisely, given a set of variables $X_1, \dots, X_m$, each variable $X_i$ can be represented in terms of
some function $F_i$ and its parents $\mathcal P(X_i)$ as
\begin{equation}\label{eq:general_scm}
  X_i = F_i\bigl(\mathcal P(X_i), N_i\bigr),
\end{equation}
where $N_i$ are independent noise terms with a given distribution.
In practice, the SCM in Eq.~\eqref{eq:general_scm} is often too general and one considers a more
restricted class.
In particular, we focus on \textit{Causal Additive Models} (CAM)~\cite{cam}, where both $F_i$ and the noise
are additive, such that (for univariate functions $f_{ij}$)
\begin{equation}
  X_i = \sum_{X_j \in \mathcal P(X_i)} f_{ij}(X_j) + N_i.
\end{equation}
A special case are linear causal models with $f_{ij}(x) = \beta_{ij}x$, s.t.\
$X_i = \sum_{X_j \in \mathcal P(X_i)} \beta_{ij} X_j + N_i$.

In the case of time series, the functions $F_i$ can in principle depend on time, which creates additional difficulties
for estimating causal relationships between variables.
Within the proposed framework, we will assume that the functions $F_i$ are invariant over time and that,
in particular, the causal dependence between variables does not change over time:
\begin{equation}
  X_i(t) = \sum_{X_j(t') \in \mathcal P(X_i(t))} f_{ij}\bigl(X_j(t')\bigr) + N_i,
\end{equation}
where necessarily $t' \leq t$ for each $X_j(t') \in P(X_i(t))$, in order to preserve the order of time.

In general, one cannot fully resolve the causal graph from observational data generated by a fully general SCM as in Eq.~\eqref{eq:general_scm}.
However, if the model is restricted to specific classes, such as a non-linear CAM, the full DAG is, in principle, identifiable~\cite{cam}.
Naturally, in a real use case the class of models must be assumed or known \textit{a priori}.
Moreover, while a linear SCM with Gaussian noise renders the DAG unidentifiable for IID data, this is not always the case for time series~\cite[Chapter 10]{peters2017elements}.

\subsection{Related work}

When novel methods are developed, the performance is usually evaluated on synthetic data
(cf.~\cite{peters2013causal}), which helps to identify strengths and weaknesses of the methods.
Unfortunately, the results cannot be directly compared to other methods, as the generated data is often
not made available.
For this reason, it is desired to create general benchmarks that can be used to compare methods.
One example is the benchmark created for the ChaLearn challenge for pairwise causal discovery~\cite{guyon2019cause}.
Recent work has been done to create a unified benchmark for causal discovery for time series.
CauseMe~\cite{runge2019inferring} contains a mix of synthetic, hybrid,
and real data based on challenges found in climate and weather data.
For these scenarios, the platform provides a ground truth
(based on domain knowledge for real data).

We see three points to how our proposed framework goes beyond the capabilities of CauseMe.
First, CauseMe is based on a ``static'' set of data used for benchmarking results.
This increases the chance of ``overfitting'' new methods to perform well on the specific use case covered in
the benchmark, rather than to perform well in general.
Our proposed framework allows one to generate vast amounts of data with different properties, including number of
observations and number of variables, enabling the user to select more robust hyperparameters that perform well under a
diverse selection of problems.
Second, our framework provides a greater flexibility to the user, which allows them to understand
the behavior of the method in specific edge cases (e.g.,\ when underlying assumptions are violated),
or how a method scales with the number of time series.
Third, the proposed framework contributes to reproducibility.
It allows the user to specify the configuration used for an experiment,
based on which others can regenerate the very same data, facilitating the reproduction of their results.

\section{Data generating process} \label{sec:data_generating_process}

We now describe the proposed data generating process. The general idea follows three steps: (i)~specify and generate a time series causal graph, (ii)~specify and generate a structural causal model (SCM), and (iii)~specify the noise and runtime configuration to generate synthetic time series data.
We expose a hierarchical \texttt{DataGenerationConfig} object, which contains subconfiguration objects for the causal graph (\texttt{CausalGraphConfig}), for the SCM (\texttt{FunctionConfig}), and to generate data (\texttt{NoiseConfig} and \texttt{RuntimeConfig}).
See~\autoref{app:example_config} for a complete example.

\begin{figure}[htb]
  \centering
  \includegraphics[width=0.985\linewidth]{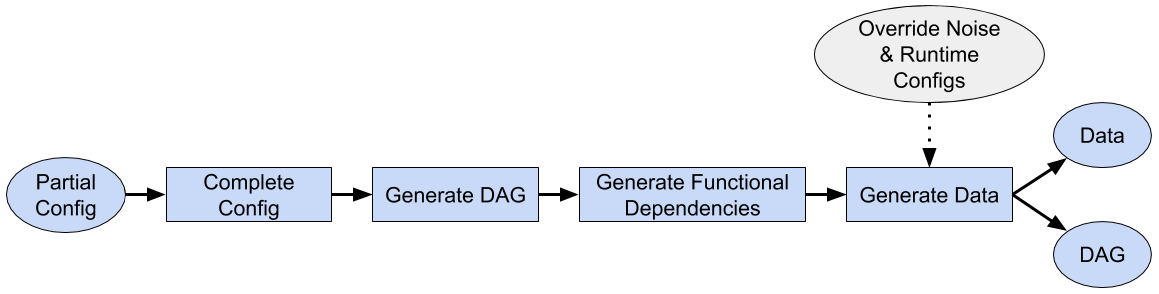}
  \caption{The user provides a \texttt{DataGenerationConfig} object. This can be partially defined and default values based on the complexity setting will be used to complete the configuration. Given the completed configuration, a time series causal graph (a Full Time Graph, cf.~\autoref{sec:brief_overview}) is randomly generated. For each edge of the DAG, a functional dependency is randomly chosen, resulting in a randomly generated SCM from which data can be generated. Multiple data sets with varying number of observations can be returned for a single SCM. Therefore, the user receives a list of data sets and a single DAG.
  Optionally, the user can override the original \texttt{NoiseConfig} and/or \texttt{RuntimeConfig} provided with the \texttt{DataGenerationConfig}. E.g., the user can change the noise distributions or regenerate a data set to also return unobserved variables.}
  \label{fig:high_level_process}
\end{figure}

\autoref{fig:high_level_process} captures a high-level overview of the process to go from a partially defined configuration to generated data, while \autoref{algo:data_generating_process}, \autoref{algo:ts_causal_graph_generating_process}, and~\autoref{algo:scm_generating_process} in~\autoref{app:algos} detail the full process. The concept of {\it complexity} exists in the configurations. Each configuration object allows the user to make the problem as complex or simple as they would like. Using the \texttt{complexity} parameter allows the system to specify default values if the user does not want to fully specify the configuration.

We expose a configuration for four types of variables: targets, features, latent, and noise. This provides the user with the capability to define the structure around specific variables. For example, when performing causal feature selection, such as in SyPI~\cite{mastakouri2020necessary}, one may want to ensure a target variable is a sink node, i.e.,\ it has no children. This is a key feature of the graph, function, and noise configurations as they allow one to fully specify the model based on the assumptions of ones' method, such as causal sufficiency and linearity for DYNOTEARS~\cite{pamfil2020dynotears}.
Additionally, sparsity of the causal graph can easily be controlled by specifying the likelihood of edges (as a universal setting or for each variable type) and the maximum number of parents and children for each variable type.

\begin{figure}[htb]
  \centering
  \subfigure[Example time series causal graph]{
    \centering
    \includegraphics[height=0.1721\textheight]{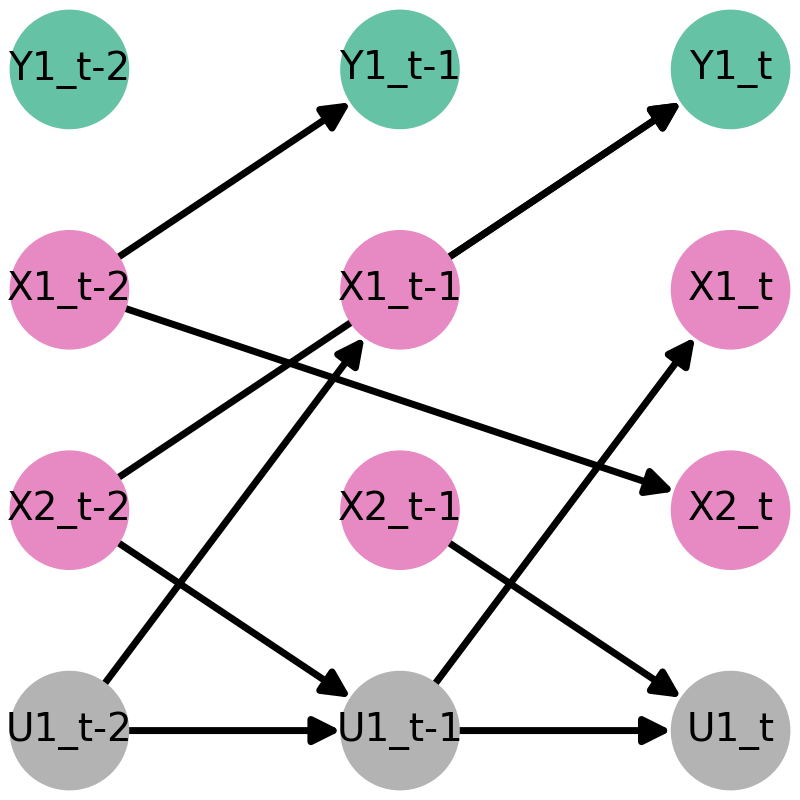}
    \label{fig:time_series_causal_graph}
  }\qquad
  \subfigure[Sample data]{
    \centering
    \includegraphics[height=0.1721\textheight]{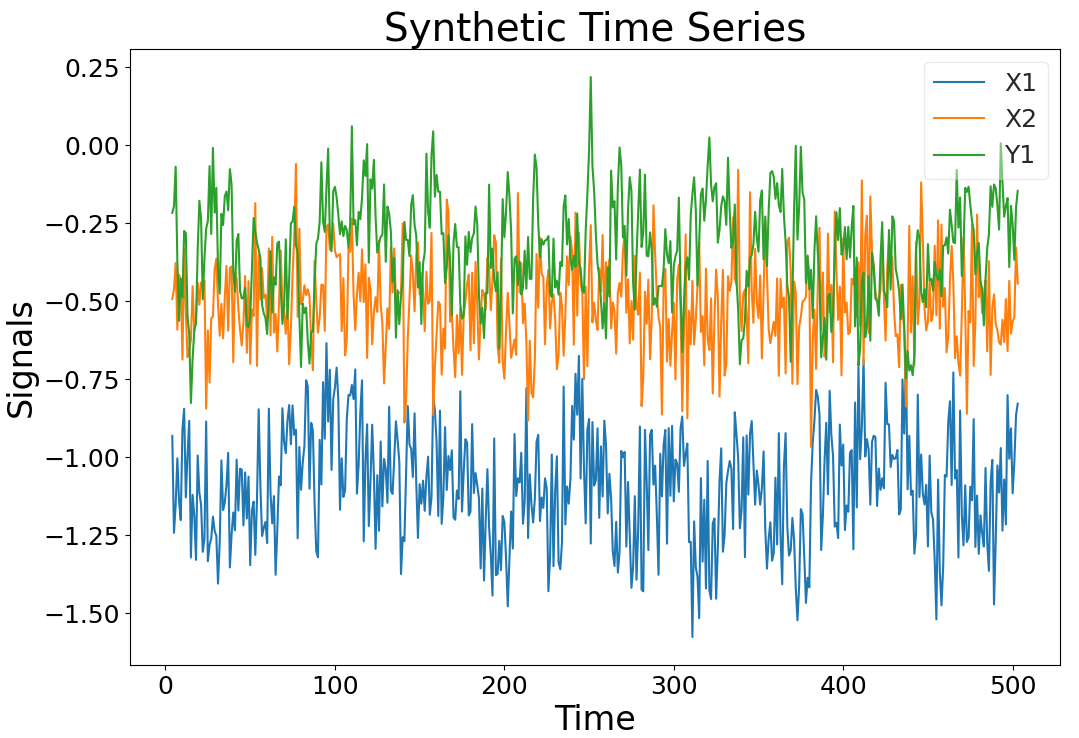}
    \label{fig:sample_data}
  }
  \caption{\subref{fig:time_series_causal_graph}~A simple time series causal graph (a Full Time Graph, cf.~\autoref{sec:brief_overview}) with a maximum lag of two, one target variable ($Y_1$), two feature variables ($X_1$ and $X_2$), and one latent variable ($U_1$); the noise nodes are not displayed for simplicity. $U_1$ shows an autoregressive relationship. \subref{fig:sample_data} The three {\it observed} time series drawn from a SCM with the causal graph in~\subref{fig:time_series_causal_graph}.
  }
  \label{fig:data_generating_example}
\end{figure}

\autoref{fig:data_generating_example}~\subref{fig:time_series_causal_graph} shows causal sufficiency being broken as we have introduced an unobserved variable. \autoref{fig:data_generating_example}~\subref{fig:sample_data} displays the time series for the three observed variables. However, if desired, it is possible to return all the synthetic data, including the latent variables and the noise variables by setting \texttt{return\_observed\_data\_only} to \texttt{False} in the \texttt{RuntimeConfig}.
Another powerful feature is defining the noise distributions after creating the SCM. The user can easily generate new data with different noise distributions and/or signal-to-noise ratio without needing to create a new SCM as depicted in~\autoref{fig:high_level_process}.
Finally, the process is open source and an example script is provided to demonstrate the effects of the various configuration settings, to provide further information on the complexity settings, and to allow users to generate data for their use cases.\footnote[1]{\url{https://github.com/causalens/cdml-neurips2020}}

\section{Experiments} \label{sec:experiments}

In order to demonstrate the breadth of data that can be created using the process proposed
in~\autoref{sec:data_generating_process},
we performed several experiments (see~\autoref{table:experiments} below and~\autoref{table:supplemental_experiments}
in \autoref{app:supplemental_results}),
each of which invalidates a specific assumption of the methods under test.
The goal is not to fully evaluate each method, but to show how the proposed data generating process supports testing.

The established methods were chosen due to their popularity and the availability of open source Python implementations.
Additionally, we evaluated a custom implementation of a multivariate version of
Granger Causality~\cite{arnold2007temporal,shojaie2010discovering}.
We also tested a bivariate version of Granger Causality~\cite{granger1969investigating} and the
PC~\cite{spirtes2000constructing} algorithm (based on PCMCI~\cite{Rungeeaau4996}).
We do not report results for these two methods, as they significantly underperformed.
Supplemental experiments and results are provided in~\autoref{app:supplemental_results}.

The methods and hyperparameters are listed in~\autoref{table:methods}.
PCMCI(+) was used with Partial Correlation and a threshold of $\alpha=0.02$ for statistical significance.
Note that the hyperparameters have been chosen manually to avoid an overly ``aggressive'' selection of links,
which would result in high FPRs or FNRs (cf.~\autoref{sec:metrics}).
We note that the results are sensitive to hyperparameter choices, generally resulting in a tradeoff between
higher TPRs and TNRs (cf.~\autoref{fig:pcmci_hyperparameters} and~\autoref{fig:dynotears_hyperparameters}).

For each experiment, 200 unique SCMs were generated from the same parameterization space defined for the
specific experiment and a single data set with 1000 samples was generated from each SCM.
For the causal sufficiency experiment, the number of feature nodes is kept at 10, while the number of
latent nodes increases from 0 to 20. For the other experiments, the parameterization space allows for a variable number
of nodes to not limit the experiments to a single graph size.\footnote[2]{Experiment data sets are provided:
\url{https://github.com/causalens/cdml-neurips2020}}
Additionally, the metrics are normalized to allow for a fair comparison between varying graph sizes.
The results presented in~\autoref{sec:results} capture the average metrics (with a maximum lag to consider of
$\ell_{\max}=5$), defined in~\autoref{sec:metrics}, for each causal discovery method across the 200 data sets with known
causal ground truth.
The synthetic data never contained true lags longer than 5 timesteps in the past.
Finally, to demonstrate the effect of the choice of hyperparameters on PCMCI and DYNOTEARS, we performed the causal
sufficiency experiment using 100 unique SCMs and a single data set with 500 samples
generated from each SCM while modifying two hyperparameters of each method.

\begin{table}[hbt]
 \caption{Experiments}
 \label{table:experiments}
 \centering
 \scalebox{0.794}{
   \begin{tabularx}{\textwidth}{lX}
     \toprule
     Name & Description \\
     \midrule
     1. Causal Sufficiency & The number of observed variables remains fixed while the number of latent variables is increased. \\
     2. Non-Linear Dependencies & The likelihood of linear functions is decreased while the likelihood of monotonic and periodic functions is increased. \\
     3. Instantaneous Effects & The minimum allowed lag in the SCM is reduced from 1 to 0. \\
     \bottomrule
   \end{tabularx}
 }
\end{table}

\begin{table}[htb]
 \caption{Causal discovery methods and their chosen hyperparameters}
 \label{table:methods}
 \centering
 \scalebox{0.794}{
   \begin{tabularx}{\textwidth}{p{0.2175\textwidth}p{0.4\textwidth}p{0.3\textwidth}}
     \toprule
     Name & Source & Hyperparameters \\
     \midrule
     PCMCI~\cite{Rungeeaau4996} & Python package \href{https://github.com/jakobrunge/tigramite}{tigramite} version 4.2 & tau\_min~=~0, tau\_max~=~5, pc\_alpha~=~0.01 \\
     PCMCI+~\cite{runge2020} & Python package \href{https://github.com/jakobrunge/tigramite}{tigramite} version 4.2 & tau\_min~=~0, tau\_max~=~5, pc\_alpha~=~0.05 \\
     DYNOTEARS~\cite{pamfil2020dynotears} & Python package \href{https://github.com/quantumblacklabs/causalnex}{causalnex} & lambda\_w~=~lambda\_a~=~0.15, w\_threshold~=~0.0,  p~=~5 \\
     VAR-LiNGAM~\cite{hyvarinen2010estimation} & Python package \href{https://github.com/cdt15/lingam}{lingam} & prune~=~True, criterion~=~`aic', lags~=~5 \\
     Multivariate Granger Causality~\cite{arnold2007temporal,shojaie2010discovering} & Cross-validated Lasso regression (\href{https://github.com/scikit-learn/scikit-learn}{scikit-learn}) + one-sided t-test (\href{https://github.com/statsmodels/statsmodels}{statsmodels}) & cv\_alphas~=~[0.02, 0.05, 0.1, 0.2, 0.3, 0.4, 0.5], max\_lag~=~5 \\
     \bottomrule
   \end{tabularx}
 }
\end{table}

\subsection{Metrics and evaluation details} \label{sec:metrics}

In order to evaluate a time series causal graph, we must define a maximum lag variable $\ell_{\max}$,
which controls the largest potential lag we want to evaluate.
Given $m$ time series $X_1, \dots, X_m$, we define the ``universe'' of variables
\begin{equation}
  \mathbb X_t := \Bigl\{X_i(t - s) ~|~  i=1, \dots, m ~\textrm{ and }~ s = 0, \dots, \ell_{\max} \Bigr\}
\end{equation}
and the set of possible links with maximal lag $\ell_{\max}$ is given by
\begin{equation}
  \label{eqn:links}
  \mathbb L_t := \Bigl\{X_i(t - s) \rightarrow X_j(t) ~|~ i, j \in \{1, \dots, m\} ~\textrm{ and }~ s = 0, \dots,
  \ell_{\max} ~\textrm{ s.t. }~ s > 0~\textrm{ or }~ i\ne j \Bigr\}.
\end{equation}
As a special case, Eq.~\eqref{eqn:links} contains the instantaneous links
\begin{equation}
  \label{eqn:instantaneous_links}
    \tilde{\mathbb L}_t := \{X_i(t) \rightarrow X_j(t) ~|~ i, j \in \{1, \dots, m\},~ i\ne j\}.
\end{equation}
Note that the latter coincides with all the possible links $m(m-1)$ prominent in an IID setup.
The total number of links is given by $|\mathbb L_t| = (\ell_{\max} + 1) m^2 - m$.
Due to the acyclicity constraint for instantaneous links $\tilde{\mathbb L}_t$, a valid DAG can contain at
most $\ell_{\max} m^2 + m(m-1)/2$ links.

One can then define the True Positives (TP) as the correctly identified links in $\mathbb L_t$, and
similarly for True Negatives (TN), False Positives (FP) and False Negatives (FN).
\autoref{table:metrics}~contains the metrics used to evaluate the performance,
including NTP, NFP and NFN, which can be used to derive SHD, F1, as well as the True Positive Rate:
TPR = NTP / (NTP + NFN).
Using these normalized values allows for a more intuitive understanding of how each component of
the SHD and F1 metrics behave.
Note that SHD = NFP + NFN, such that a SHD of 5\% means that the method misclassified 5\% of the edges.
For a typical graph we have much more Negatives (N) than Positives (P), such that the NTP and NFN will
be on a much smaller scale than TPR and FNR, making a direct comparison impossible.
However, NFP and False Positive Rate (FPR) should be roughly on the same scale.

\begin{table}[htb]
 \caption{Metrics}
 \label{table:metrics}
 \centering
 \scalebox{0.794}{
   \begin{tabularx}{\textwidth}{lcX}
     \toprule
     Name & Acronym & Description \\
     \midrule
     F1-Score & F1 & Calculated as F1 = TP / (TP + (FP + FN) / 2). \\
     Structural Hamming Distance~\cite{tsamardinos2006max} & SHD & Normalized as (FP + FN) / $|\mathbb L_t|$. \\
     Normalized True Positives & NTP & Calculated as TP / $|\mathbb L_t|$. \\
     Normalized False Positives & NFP & Calculated as FP / $|\mathbb L_t|$. \\
     Normalized False Negatives & NFN & Calculated as FN / $|\mathbb L_t|$. \\
     \bottomrule
   \end{tabularx}
 }
\end{table}

\subsection{Results} \label{sec:results}

\begin{figure}
  \centering
  \subfigure[F1]{
    \centering
    \includegraphics[width=0.46\linewidth]{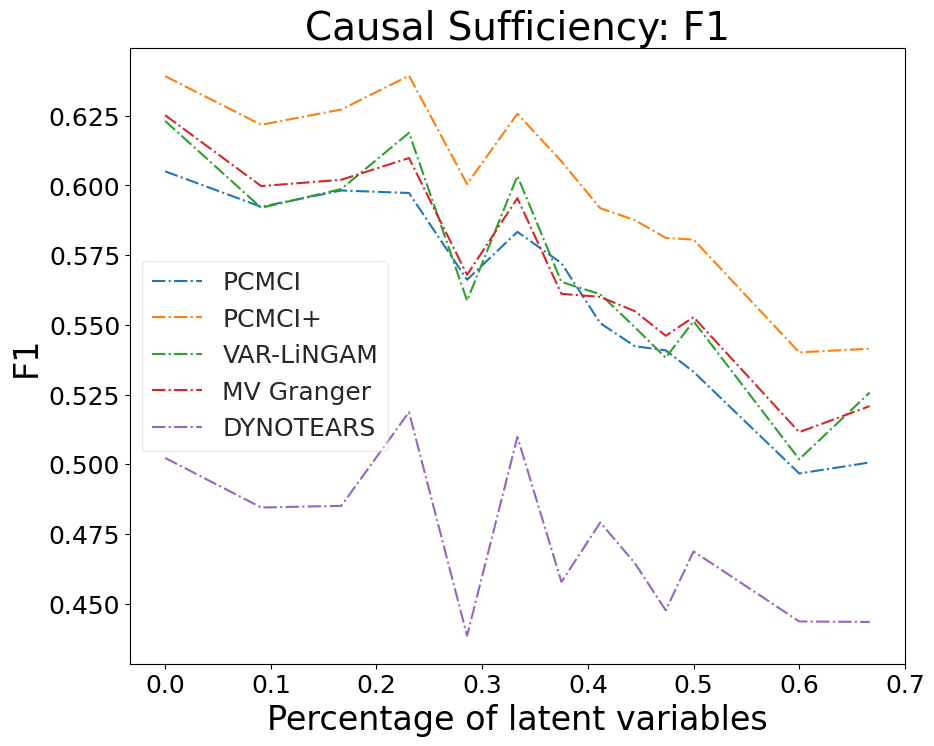}
    \label{fig:f1_causal_sufficiency}
  }\hfill
  \subfigure[SHD]{
    \centering
    \includegraphics[width=0.46\linewidth]{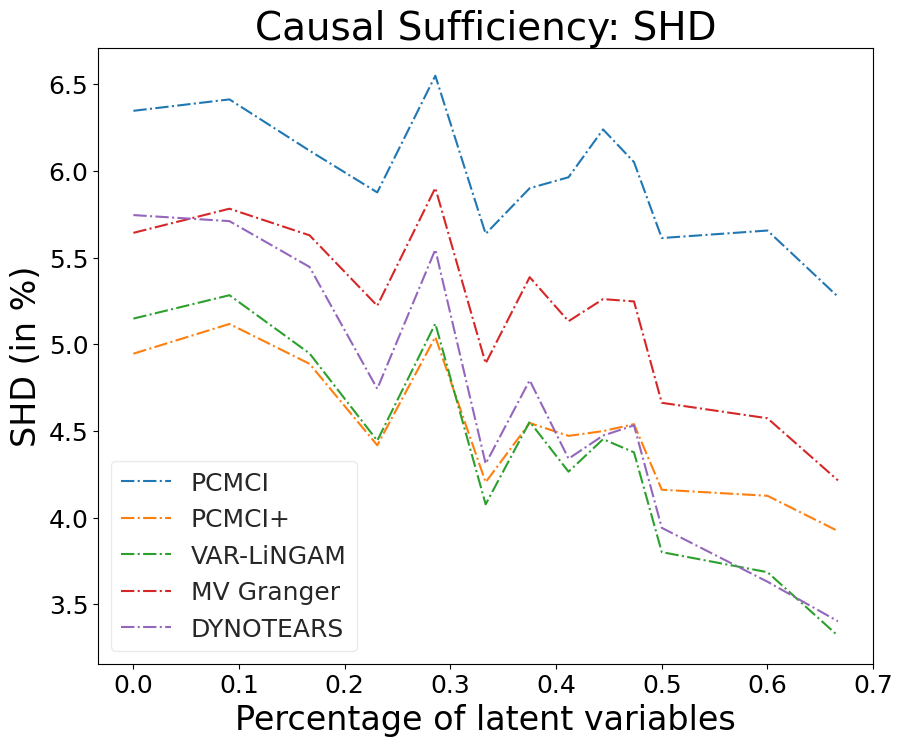}
    \label{fig:shd_causal_suffiency}
  }
  \subfigure[Normalized true positives]{
    \centering
    \includegraphics[width=0.31\linewidth]{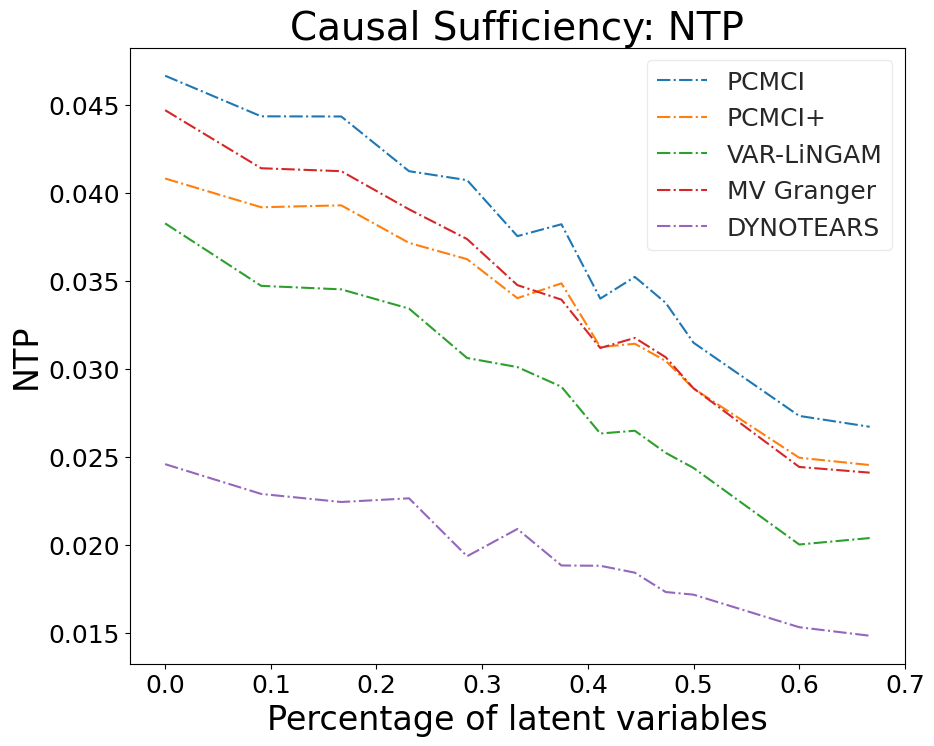}
    \label{fig:ntp_causal_suffiency}
  }\hfill
  \subfigure[Normalized false positives]{
    \centering
    \includegraphics[width=0.31\linewidth]{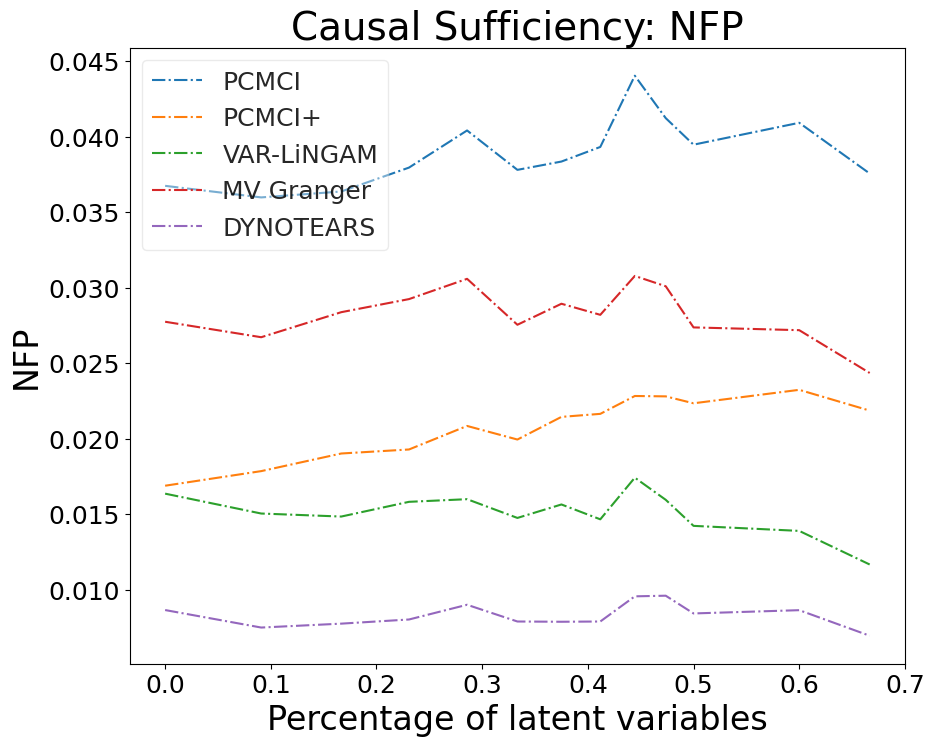}
    \label{fig:nfp_causal_suffiency}
  }\hfill
  \subfigure[Normalized false negatives]{
    \centering
    \includegraphics[width=0.31\linewidth]{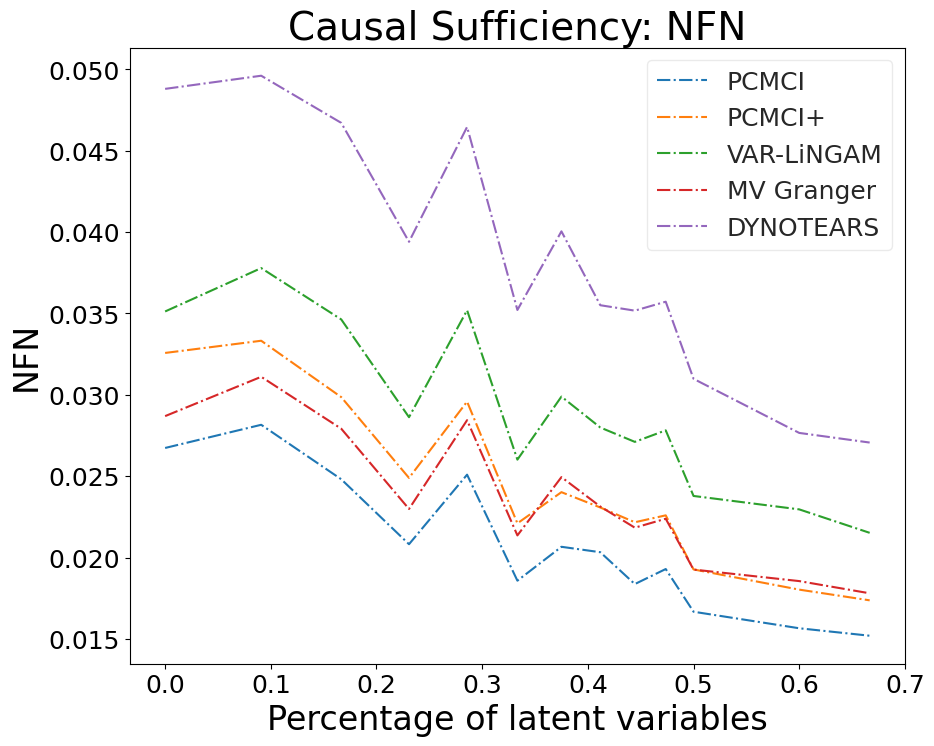}
    \label{fig:nfn_causal_suffiency}
  }
  \caption{Causal Sufficiency - The percentage of latent variables is the number of latents divided by the number of observed plus latent. \subref{fig:f1_causal_sufficiency}~F1 decreases for all methods as more latent variables are added. \subref{fig:shd_causal_suffiency}~Note that SHD also decreases. As defined in~\autoref{table:metrics}, SHD is only a function of FPs and FNs, while F1 is also a function of TPs. \subref{fig:ntp_causal_suffiency} and~\subref{fig:nfn_causal_suffiency} respectively show that TPs and FPs decrease at a similar rate as more latent variables are added. TPs have a larger effect on F1, hence why we observe an overall decrease. \subref{fig:nfp_causal_suffiency} The relative minor changes in FPs when compared to the larger decrease in FNs leads to an overall decrease in SHD.}
  \label{fig:causal_sufficiency}
\end{figure}

\begin{figure}
  \centering
  \subfigure[F1]{
    \centering
    \includegraphics[width=0.46\linewidth]{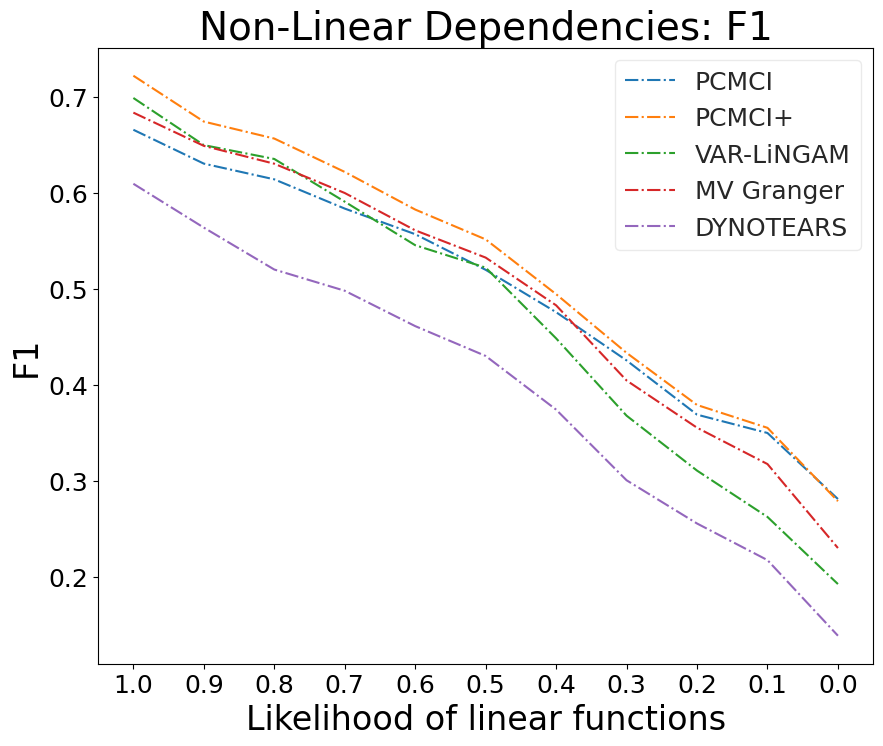}
    \label{fig:f1_nonlinear_dependecies}
  }\hfill
  \subfigure[SHD]{
    \centering
    \includegraphics[width=0.46\linewidth]{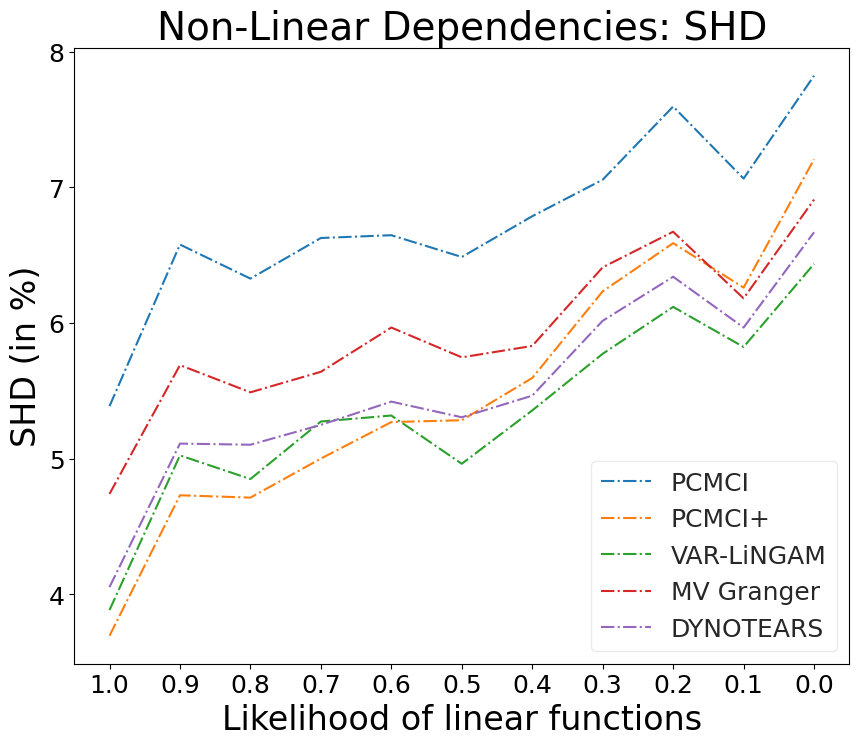}
    \label{fig:shd_nonlinear_dependecies}
  }
  \caption{Non-Linear (Monotonic and Periodic) Dependencies - \subref{fig:f1_nonlinear_dependecies}~F1 decreases and~\subref{fig:shd_nonlinear_dependecies} SHD increases as the percentage of linear dependencies in the system is decreased. A key observation is that VAR-LiNGAM, multivariate Granger, and DYNOTEARS have the largest decreases in F1, which is expected as they are linear vector autoregressive models. PCMCI and PCMCI+ are negatively impacted by the use of Pearson correlation within the conditional independence test.
  We explicitly apply linear methods to a non-linear setting to demonstrate how the methods perform when this assumption is violated; we are not expecting linear methods to perform well for non-linear problems.
  }
  \label{fig:nonlinear_dependecies}
\end{figure}

\begin{figure}
  \centering
  \subfigure[F1]{
    \centering
    \includegraphics[width=0.4125\linewidth]{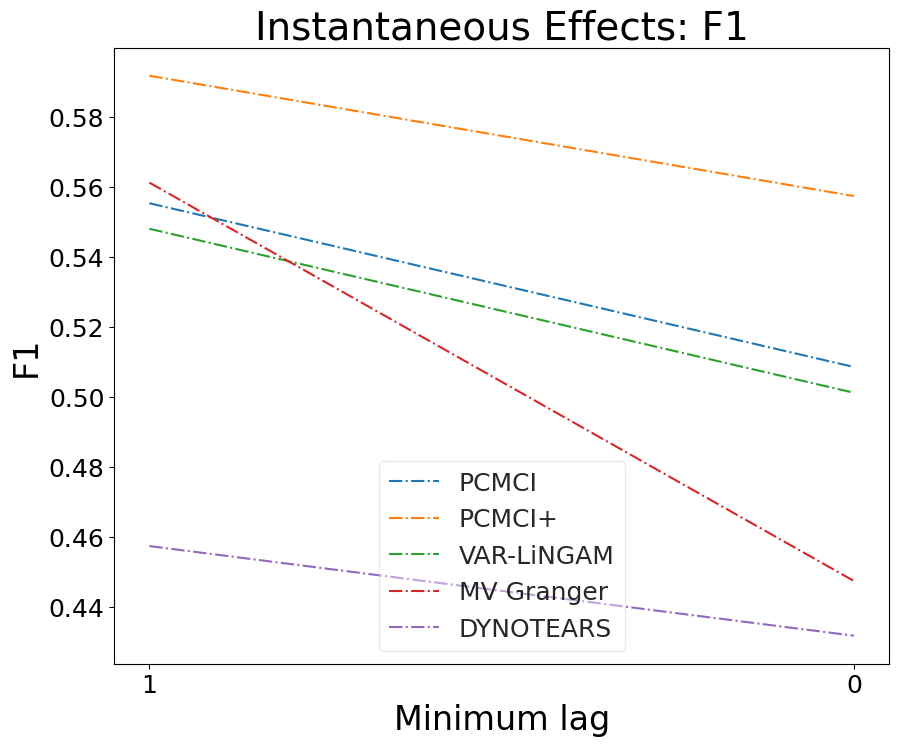}
    \label{fig:f1_instantaneous_effects}
  }\qquad
  \subfigure[SHD]{
    \centering
    \includegraphics[width=0.4125\linewidth]{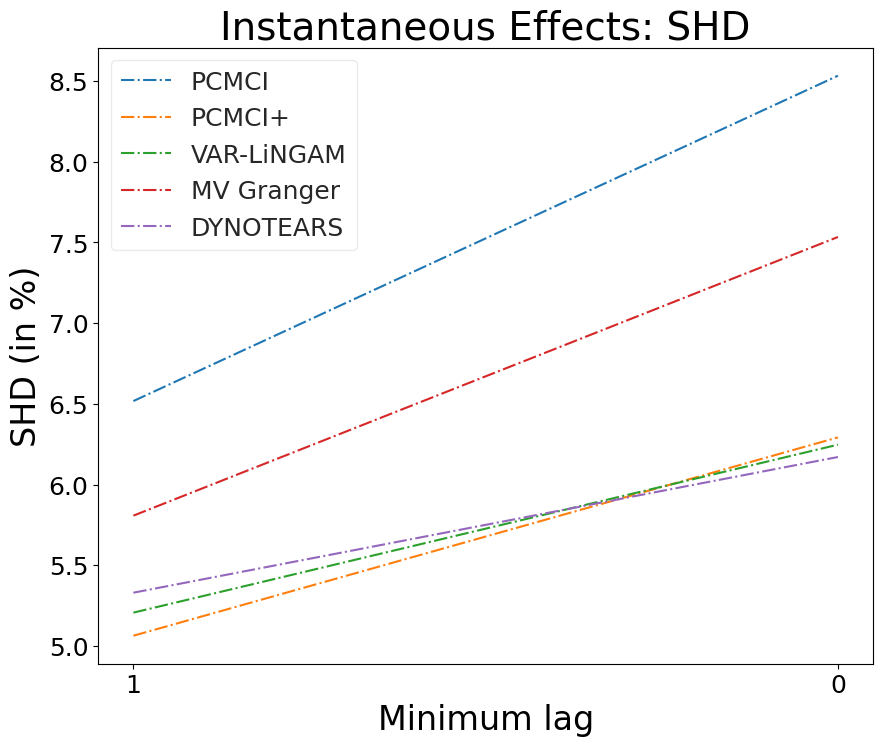}
    \label{fig:shd_instantaneous_effects}
  }\\
  \subfigure[Normalized true positives]{
    \centering
    \includegraphics[width=0.4125\linewidth]{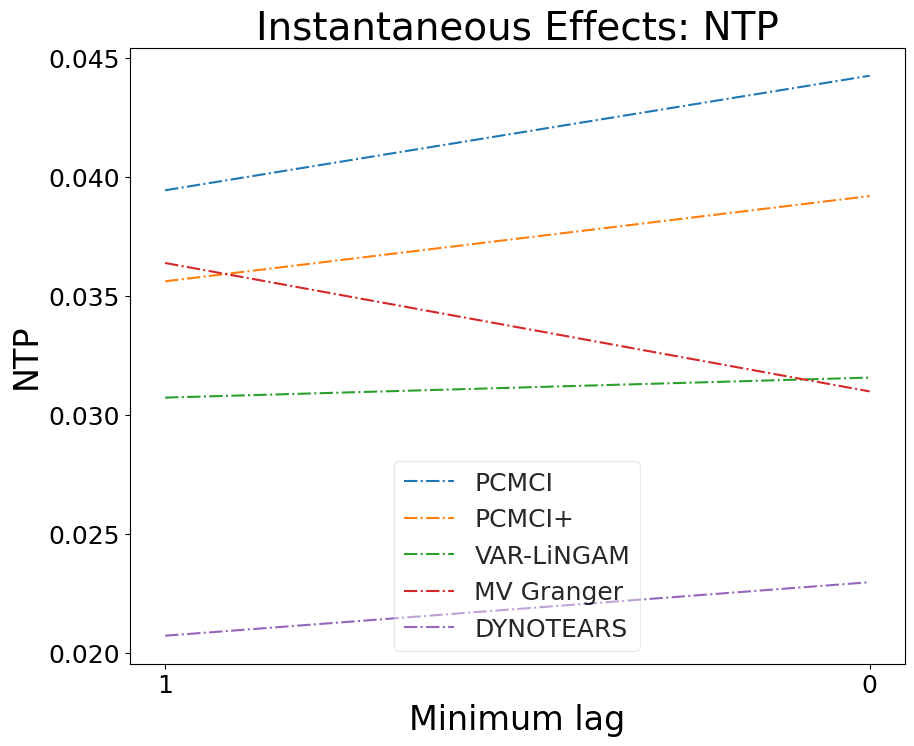}
    \label{fig:ntp_instantaneous_effects}
  }\qquad
  \subfigure[Normalized false positives]{
    \centering
    \includegraphics[width=0.4125\linewidth]{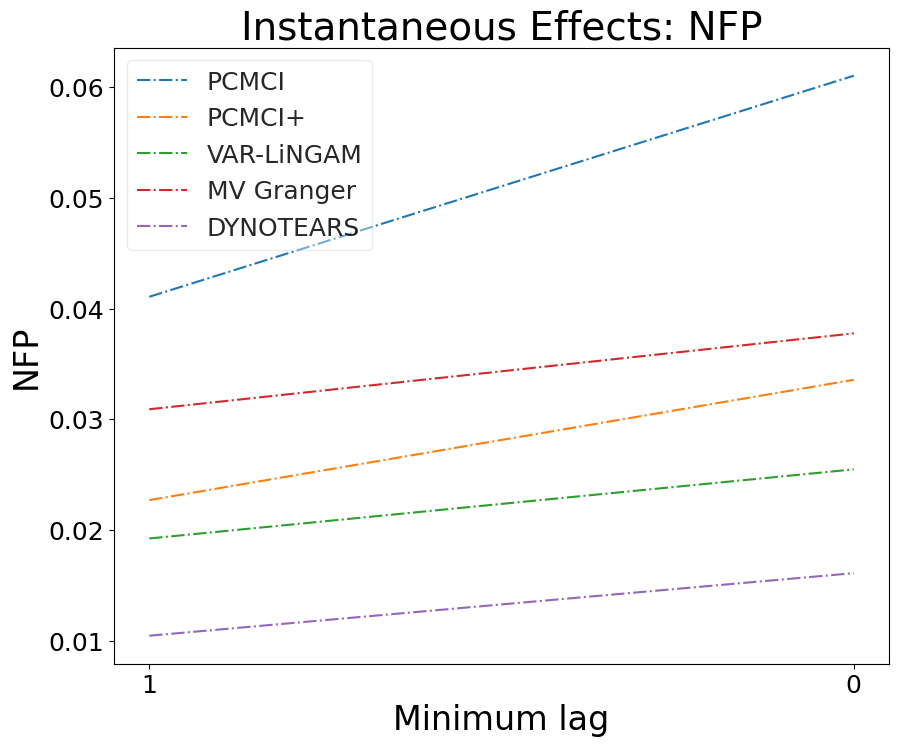}
    \label{fig:nfp_instantaneous_effects}
  }
  \caption{Instantaneous Effects - \subref{fig:f1_instantaneous_effects}~F1 decreases and~\subref{fig:shd_instantaneous_effects} SHD increases for all methods when instantaneous effects are allowed.
  By construction, multivariate Granger does not regress on covariates with lag 0 and hence cannot identify instantaneous effects; the TPs~\subref{fig:ntp_instantaneous_effects} drop substantially. PCMCI+ is specifically designed to handle instantaneous links~\cite{runge2020}. PCMCI is capable of returning edges at lag 0, but the edges are not oriented. As such, for each TP, there will be a FP, as seen by the steep increase in FPs~\subref{fig:nfp_instantaneous_effects} for PCMCI when compared to the slower increase for PCMCI+.
  Note that this is one scenario when PCMCI does not return a valid DAG, but a CPDAG (cf.~\autoref{sec:background}).}
  \label{fig:instantaneous_effects}
\end{figure}

\begin{figure}
  \centering
  \subfigure[F1]{
    \centering
    \includegraphics[width=0.48725\linewidth]{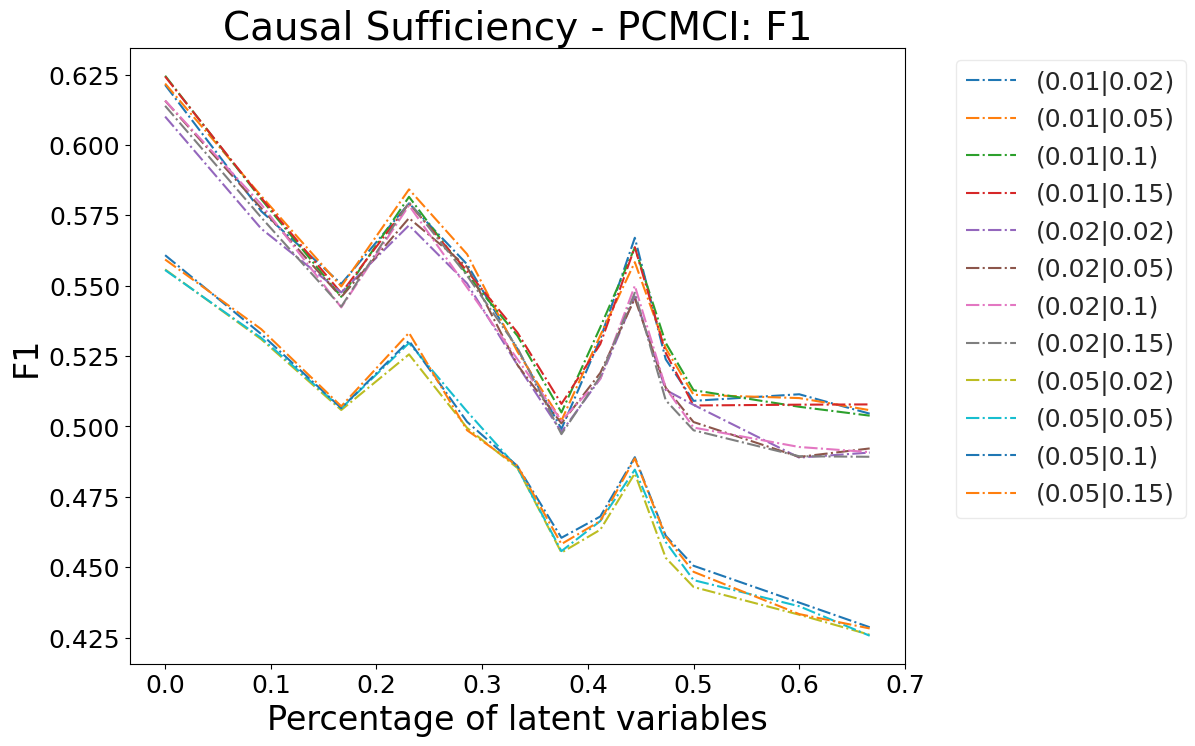}
    \label{fig:f1_pcmci_hyperparameters}
  }\hfill
  \subfigure[SHD]{
    \centering
    \includegraphics[width=0.48725\linewidth]{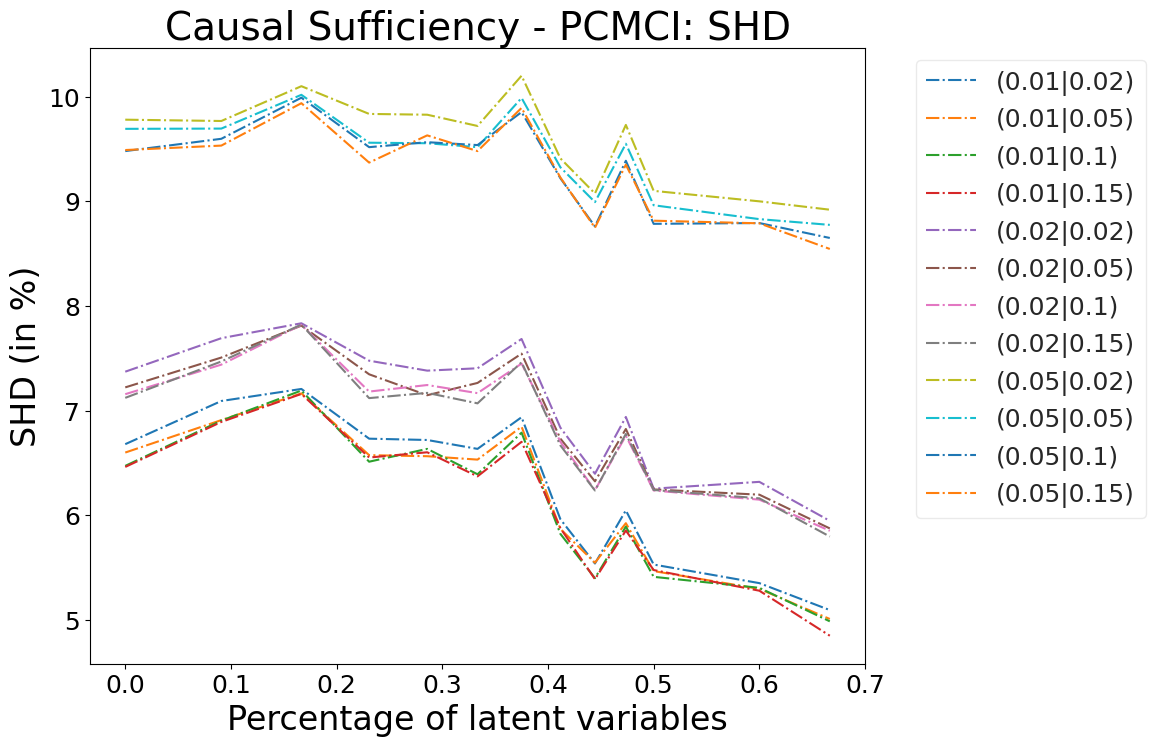}
    \label{fig:shd_pcmci_hyperparameters}
  }
  \caption{
  \subref{fig:f1_pcmci_hyperparameters}~F1 and \subref{fig:shd_pcmci_hyperparameters}~SHD for different
  choices of hyperparameters for PCMCI.
  We varied the p-value threshold for the final selection (first parameter in the legend) over \{0.01, 0.02, 0.05\}
  and the p-value for the PC algorithm (second parameter in the legend) over \{0.02, 0.05, 0.1, 0.15\}.
  Both F1 and SHD show that the variation of the p-value for the PC step only has a minor
  effect on the performance, and the majority is accounted for by the first parameter, which is the
  p-value threshold used for the final score.
  From the results, the best performance is achieved for the final p-value threshold equal to 0.01 and the
  p-value for the PC step in the range of [0.02, 0.2].
  }
  \label{fig:pcmci_hyperparameters}
\end{figure}

\begin{figure}
  \centering
  \subfigure[F1]{
    \centering
    \includegraphics[width=0.48725\linewidth]{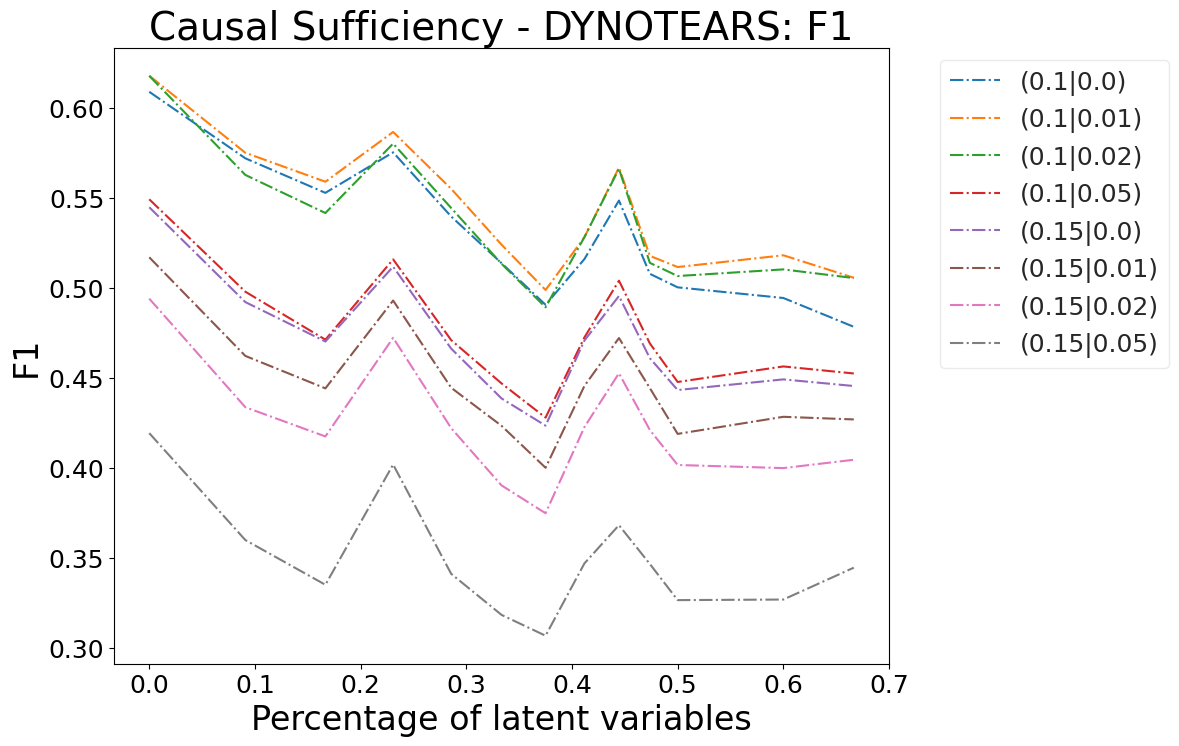}
    \label{fig:f1_dynotears_hyperparameters}
  }\hfill
  \subfigure[SHD]{
    \centering
    \includegraphics[width=0.48725\linewidth]{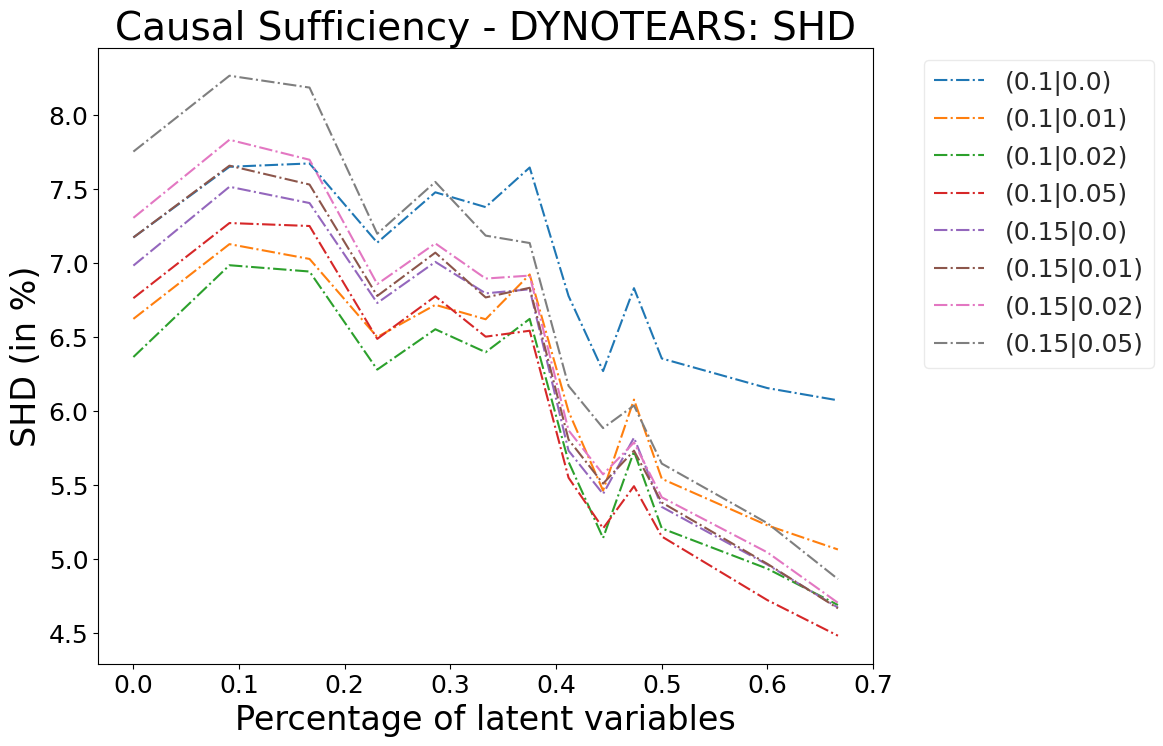}
    \label{fig:shd_dynotears_hyperparameters}
  }
  \caption{
  \subref{fig:f1_dynotears_hyperparameters}~F1 and \subref{fig:shd_dynotears_hyperparameters}~SHD for different
  choices of hyperparameters for DYNOTEARS.
  The first parameter in the legend is equal to the L1-penalty on all the variables (specified by
  lambda\_w and lambda\_a, which are here chosen to be equal).
  The second parameter is w\_threshold, which is used as a threshold for the cutoff of weights.
  The results suggest that the optimal choices are moderate values of the L1-penalties
  (lambda\_w and lambda\_a between 0.5 and 1.0), combined with a small value for w\_threshold.
  }
  \label{fig:dynotears_hyperparameters}
\end{figure}

Performance of the causal discovery methods under evaluation against the metrics defined
in~\autoref{sec:metrics} are provided in \autoref{fig:causal_sufficiency}, \autoref{fig:nonlinear_dependecies},
and~\autoref{fig:instantaneous_effects} for the experiments defined in~\autoref{table:experiments}, respectively.
\autoref{fig:causal_sufficiency} and~\autoref{fig:nonlinear_dependecies} show how all the methods are affected
by latent variables and non-linear dependencies, respectively.
The compared methods share the assumptions of causal sufficiency and linearity, and, as expected, their
performance degrades with the violation of these assumptions.
The introduction of non-linearities decreases the methods' performance to a much larger extent than the
invalidation of causal sufficiency.
However, the choice of hyperparameters has as much of an effect, even when assumptions are met.
Not all hyperparameters have the same importance and, accordingly, we suggest a range of values
in~\autoref{fig:pcmci_hyperparameters} and~\autoref{fig:dynotears_hyperparameters} for PCMCI and DYNOTEARS,
respectively. Finally, results for the supplemental experiments are provided in~\autoref{app:supplemental_results}.

\section{Conclusion} \label{sec:conclusion}

We have proposed a process to generate synthetic time series data with a known ground truth causal structure
and demonstrated its functionality by evaluating prominent causal discovery techniques for time series data.
The process is easily parameterizable yet provides the capability to generate data from vastly different scenarios.
The process is open source and an example script to allow users to generate data for their use cases
is provided.
This is the main contribution over existing benchmarks, such as CauseMe~\cite{runge2019inferring}, as their scope
is restricted with respect to number of observations, number of variables, and specific dynamic system challenges.
We have demonstrated how the proposed framework can be used to discriminate the performance of different
causal discovery methods under a variety of conditions and how this framework can be used for fine tuning
hyperparameters without the fear of overfitting to a target benchmark.

We believe our proposed data generating process can be used to support both research and the practical applications of
causal discovery.
As a result of our experiments, we have identified two important research directions: (i) the continued development
of efficient, non-linear methods, and (ii) less reliance on hyperparameters.
To address sparsity of data and lack of ground truth, a researcher/practitioner will generate synthetic data
in agreement with their domain knowledge.
The resulting synthetic benchmark will provide a principled approach to the
development/selection of a method and its hyperparameters, as opposed to simply overfitting to one's data.

\paragraph{Future work}
The current implementation can be extended in various directions, particularly around functional forms, noise,
and dynamic graphs.
Currently, the process only supports additive, homoscedastic noise;
adding support for multiplicative and heteroskedastic noise would be beneficial.
The current process also produces static models.
Allowing for the distributions, function parameters, and causal graph to change over time will produce synthetic
data with changepoints, regime shifts, and/or interventions.

\begin{ack}

The authors would like to thank Microsoft for Startups for supporting this research through their
contribution of GitHub and Azure credits. We would also like to thank the two anonymous reviewers whose comments
helped us to improve and clarify the paper.

\end{ack}

\clearpage
\bibliography{ref}

\begin{thebibliography}{10}

\bibitem{arnold2007temporal}
Andrew Arnold, Yan Liu, and Naoki Abe.
\newblock Temporal causal modeling with graphical granger methods.
\newblock In {\em Proceedings of the 13th ACM SIGKDD international conference
  on Knowledge discovery and data mining}, pages 66--75, 2007.

\bibitem{barnett2009granger}
Lionel Barnett, Adam~B Barrett, and Anil~K Seth.
\newblock Granger causality and transfer entropy are equivalent for gaussian
  variables.
\newblock {\em Physical review letters}, 103(23):238701, 2009.

\bibitem{cam}
Peter B{\"u}hlmann, Jonas Peters, Jan Ernest, et~al.
\newblock {CAM}: Causal additive models, high-dimensional order search and
  penalized regression.
\newblock {\em The Annals of Statistics}, 42(6):2526--2556, 2014.

\bibitem{chickering2002optimal}
David~Maxwell Chickering.
\newblock Optimal structure identification with greedy search.
\newblock {\em Journal of machine learning research}, 3(Nov):507--554, 2002.

\bibitem{colombo2012learning}
Diego Colombo, Marloes~H Maathuis, Markus Kalisch, and Thomas~S Richardson.
\newblock Learning high-dimensional directed acyclic graphs with latent and
  selection variables.
\newblock {\em The Annals of Statistics}, pages 294--321, 2012.

\bibitem{entner2010causal}
Doris Entner and Patrik~O Hoyer.
\newblock On causal discovery from time series data using fci.
\newblock {\em Probabilistic graphical models}, pages 121--128, 2010.

\bibitem{friedman2001elements}
Jerome Friedman, Trevor Hastie, and Robert Tibshirani.
\newblock {\em The elements of statistical learning}, volume~1.
\newblock Springer series in statistics New York, 2001.

\bibitem{gerhardus2020high}
Andreas Gerhardus and Jakob Runge.
\newblock High-recall causal discovery for autocorrelated time series with
  latent confounders.
\newblock {\em arXiv preprint arXiv:2007.01884}, 2020.

\bibitem{glymour2019review}
Clark Glymour, Kun Zhang, and Peter Spirtes.
\newblock Review of causal discovery methods based on graphical models.
\newblock {\em Frontiers in genetics}, 10:524, 2019.

\bibitem{granger1969investigating}
Clive~WJ Granger.
\newblock Investigating causal relations by econometric models and
  cross-spectral methods.
\newblock {\em Econometrica: journal of the Econometric Society}, pages
  424--438, 1969.

\bibitem{guyon2019cause}
Isabelle Guyon, Alexander Statnikov, and Berna~Bakir Batu.
\newblock {\em Cause Effect Pairs in Machine Learning}.
\newblock Springer, 2019.

\bibitem{huang2020causal}
Biwei Huang, Kun Zhang, Jiji Zhang, Joseph Ramsey, Ruben Sanchez-Romero, Clark
  Glymour, and Bernhard Sch{\"o}lkopf.
\newblock Causal discovery from heterogeneous/nonstationary data.
\newblock {\em Journal of Machine Learning Research}, 21(89):1--53, 2020.

\bibitem{hyvarinen2010estimation}
Aapo Hyv{\"a}rinen, Kun Zhang, Shohei Shimizu, and Patrik~O Hoyer.
\newblock Estimation of a structural vector autoregression model using
  non-gaussianity.
\newblock {\em Journal of Machine Learning Research}, 11(5), 2010.

\bibitem{krizhevsky2009learning}
Alex Krizhevsky, Geoffrey Hinton, et~al.
\newblock Learning multiple layers of features from tiny images.
\newblock 2009.

\bibitem{lecun1998gradient}
Yann LeCun, L{\'e}on Bottou, Yoshua Bengio, and Patrick Haffner.
\newblock Gradient-based learning applied to document recognition.
\newblock {\em Proceedings of the IEEE}, 86(11):2278--2324, 1998.

\bibitem{lecun2010mnist}
Yann LeCun, Corinna Cortes, and CJ~Burges.
\newblock {MNIST} handwritten digit database.
\newblock {\em ATT Labs [Online]. Available: http://yann.lecun.com/exdb/mnist},
  2, 2010.

\bibitem{malinsky2019learning}
Daniel Malinsky and Peter Spirtes.
\newblock Learning the structure of a nonstationary vector autoregression.
\newblock {\em Proceedings of machine learning research}, 89:2986, 2019.

\bibitem{marinazzo2008kernel}
Daniele Marinazzo, Mario Pellicoro, and Sebastiano Stramaglia.
\newblock Kernel-granger causality and the analysis of dynamical networks.
\newblock {\em Physical review E}, 77(5):056215, 2008.

\bibitem{mastakouri2020necessary}
Atalanti~A Mastakouri, Bernhard Sch{\"o}lkopf, and Dominik Janzing.
\newblock Necessary and sufficient conditions for causal feature selection in
  time series with latent common causes.
\newblock {\em arXiv preprint arXiv:2005.08543}, 2020.

\bibitem{meek1997graphical}
Christopher Meek.
\newblock {\em Graphical Models: Selecting causal and statistical models}.
\newblock PhD thesis, PhD thesis, Carnegie Mellon University, 1997.

\bibitem{ng2020role}
Ignavier Ng, AmirEmad Ghassami, and Kun Zhang.
\newblock On the role of sparsity and dag constraints for learning linear dags.
\newblock {\em arXiv preprint arXiv:2006.10201}, 2020.

\bibitem{pamfil2020dynotears}
Roxana Pamfil, Nisara Sriwattanaworachai, Shaan Desai, Philip Pilgerstorfer,
  Konstantinos Georgatzis, Paul Beaumont, and Bryon Aragam.
\newblock Dynotears: Structure learning from time-series data.
\newblock In {\em International Conference on Artificial Intelligence and
  Statistics}, pages 1595--1605, 2020.

\bibitem{pearl2009causal}
Judea Pearl et~al.
\newblock Causal inference in statistics: An overview.
\newblock {\em Statistics surveys}, 3:96--146, 2009.

\bibitem{peters2013causal}
Jonas Peters, Dominik Janzing, and Bernhard Sch{\"o}lkopf.
\newblock Causal inference on time series using restricted structural equation
  models.
\newblock In {\em Advances in Neural Information Processing Systems}, pages
  154--162, 2013.

\bibitem{peters2017elements}
Jonas Peters, Dominik Janzing, and Bernhard Sch{\"o}lkopf.
\newblock {\em Elements of causal inference}.
\newblock The MIT Press, 2017.

\bibitem{runge2018causal}
Jakob Runge.
\newblock Causal network reconstruction from time series: From theoretical
  assumptions to practical estimation.
\newblock {\em Chaos: An Interdisciplinary Journal of Nonlinear Science},
  28(7):075310, 2018.

\bibitem{runge2020}
Jakob Runge.
\newblock Discovering contemporaneous and lagged causal relations in
  autocorrelated nonlinear time series datasets.
\newblock In {\em Proceedings of the Thirty-Sixth Conference on Uncertainty in
  Artificial Intelligence ({UAI})}, pages 1388--1397. {AUAI} Press, 03--06 Aug
  2020.

\bibitem{runge2019inferring}
Jakob Runge, Sebastian Bathiany, Erik Bollt, Gustau Camps-Valls, Dim Coumou,
  Ethan Deyle, Clark Glymour, Marlene Kretschmer, Miguel~D Mahecha, Jordi
  Mu{\~n}oz-Mar{\'\i}, et~al.
\newblock Inferring causation from time series in earth system sciences.
\newblock {\em Nature communications}, 10(1):1--13, 2019.

\bibitem{Rungeeaau4996}
Jakob Runge, Peer Nowack, Marlene Kretschmer, Seth Flaxman, and Dino
  Sejdinovic.
\newblock Detecting and quantifying causal associations in large nonlinear time
  series datasets.
\newblock {\em Science Advances}, 5(11), 2019.

\bibitem{schreiber2000measuring}
Thomas Schreiber.
\newblock Measuring information transfer.
\newblock {\em Physical review letters}, 85(2):461, 2000.

\bibitem{shimizu2006linear}
Shohei Shimizu, Patrik~O Hoyer, Aapo Hyv{\"a}rinen, and Antti Kerminen.
\newblock A linear non-gaussian acyclic model for causal discovery.
\newblock {\em Journal of Machine Learning Research}, 7(Oct):2003--2030, 2006.

\bibitem{shimizu2011directlingam}
Shohei Shimizu, Takanori Inazumi, Yasuhiro Sogawa, Aapo Hyv{\"a}rinen,
  Yoshinobu Kawahara, Takashi Washio, Patrik~O Hoyer, and Kenneth Bollen.
\newblock Directlingam: A direct method for learning a linear non-gaussian
  structural equation model.
\newblock {\em The Journal of Machine Learning Research}, 12:1225--1248, 2011.

\bibitem{shojaie2010discovering}
Ali Shojaie and George Michailidis.
\newblock Discovering graphical granger causality using the truncating lasso
  penalty.
\newblock {\em Bioinformatics}, 26(18):i517--i523, 2010.

\bibitem{spirtes2000constructing}
Pater Spirtes, Clark Glymour, Richard Scheines, Stuart Kauffman, Valerio
  Aimale, and Frank Wimberly.
\newblock Constructing bayesian network models of gene expression networks from
  microarray data.
\newblock 2000.

\bibitem{spirtes2000causation}
Peter Spirtes, Clark~N Glymour, Richard Scheines, and David Heckerman.
\newblock {\em Causation, prediction, and search}.
\newblock MIT press, 2000.

\bibitem{tsamardinos2006max}
Ioannis Tsamardinos, Laura~E Brown, and Constantin~F Aliferis.
\newblock The max-min hill-climbing bayesian network structure learning
  algorithm.
\newblock {\em Machine learning}, 65(1):31--78, 2006.

\bibitem{varando2020learning}
Gherardo Varando.
\newblock Learning dags without imposing acyclicity.
\newblock {\em arXiv preprint arXiv:2006.03005}, 2020.

\bibitem{weichwald2020causal}
Sebastian Weichwald, Martin~E. Jakobsen, Phillip~B. Mogensen, Lasse Petersen,
  Nikolaj Thams, and Gherardo Varando.
\newblock Causal structure learning from time series: Large regression
  coefficients may predict causal links better in practice than small p-values.
\newblock volume 123 of {\em Proceedings of the NeurIPS 2019 Competition and
  Demonstration Track, Proceedings of Machine Learning Research}, pages 27--36.
  PMLR, 2020.

\bibitem{yu2019dag}
Yue Yu, Jie Chen, Tian Gao, and Mo~Yu.
\newblock Dag-gnn: Dag structure learning with graph neural networks.
\newblock {\em arXiv preprint arXiv:1904.10098}, 2019.

\bibitem{zheng2018dags}
Xun Zheng, Bryon Aragam, Pradeep~K Ravikumar, and Eric~P Xing.
\newblock Dags with no tears: Continuous optimization for structure learning.
\newblock In {\em Advances in Neural Information Processing Systems}, pages
  9472--9483, 2018.

\bibitem{zheng2020learning}
Xun Zheng, Chen Dan, Bryon Aragam, Pradeep Ravikumar, and Eric Xing.
\newblock Learning sparse nonparametric dags.
\newblock In {\em International Conference on Artificial Intelligence and
  Statistics}, pages 3414--3425. PMLR, 2020.

\end{thebibliography}

\clearpage
\begin{appendices}

  \section{Algorithmic representation of data generating process} \label{app:algos}

  \autoref{algo:data_generating_process}, \autoref{algo:ts_causal_graph_generating_process}, and~\autoref{algo:scm_generating_process} define the data generating process proposed in~\autoref{sec:data_generating_process} and provide the low-level steps to go from a configuration to generated data as shown in \autoref{fig:high_level_process}.

  \begin{algorithm}[h]
  \SetAlgoLined
  \KwIn{\texttt{config}: DataGenerationConfig}
   Complete missing values with defaults for \texttt{config.noise\_config} (based on complexity value) and \texttt{config.runtime\_config}. \\
   Generate TimeSeriesCausalGraph per~\autoref{algo:ts_causal_graph_generating_process}. \\
   Generate StructuralCausalModel per~\autoref{algo:scm_generating_process}. \\
   \ForEach{\texttt{num\_samples} and \texttt{data\_generating\_seed} in \texttt{config.runtime\_config}}{
    Seed process using \texttt{data\_generating\_seed}. \\
    Initialize all data with zeroes. \\
    \ForEach{Noise variable $N_i$ in StructuralCausalModel}{
      \eIf{\texttt{noise\_config.noise\_variance} is provided as a range}{
        \texttt{noise\_var} $\sim$ Uniform(\texttt{noise\_config.noise\_variance}) \\
      }{
        \texttt{noise\_var} ${\leftarrow}$ \texttt{noise\_config.noise\_variance} \\
      }
      Randomly sample noise distribution from \texttt{noise\_config.distributions} with probabilities defined in \texttt{noise\_config.prob\_distributions}. \\
      $N_i \leftarrow$ Sample \texttt{num\_samples} IID samples from the chosen distribution with variance \texttt{noise\_var}. \\
    }
    \ForEach{Noise variable $N_i$ with an autoregressive edge in StructuralCausalModel}{
      \For{$t \leftarrow 1$ to \texttt{num\_samples}}{
        $N_i(t) \leftarrow N_i(t) + f_i(N_i(t-1))$ \\
      }
    }
    \For{$t \leftarrow$ \texttt{config.graph\_config.max\_lag} to \texttt{num\_samples}}{
      \ForEach{Non-noise variable $X_i$ in topological order of graph}{
        \ForEach{Parent variable $X_j$, lagged time index $t'$, and functional dependency $f_{ij}$}{
          $X_i(t) \leftarrow X_i(t) + f_{ij}\bigl(X_j(t')\bigr)$ \\
          Note: This includes the additive noise as the noise is a parent and its functional dependency is the identity function as described in~\autoref{algo:scm_generating_process}. \\
        }
      }
    }
   }
   \KwOut{Tuple[List[Dataset], TimeSeriesCausalGraph]}
   \caption{Time Series Data Generation}
   \label{algo:data_generating_process}
  \end{algorithm}

  \begin{algorithm}[h]
  \SetAlgoLined
  \KwIn{\texttt{graph\_config}: CausalGraphConfig}
   Complete missing values with defaults based on complexity value for \texttt{graph\_config}. \\
   Initialize empty DAG with $M$ nodes, where $M$ = (1 + \texttt{graph\_config.include\_noise}) $*$ (\texttt{graph\_config.num\_targets} + \texttt{graph\_config.num\_features} + \texttt{graph\_config.num\_latent}) $*$ (1 + \texttt{graph\_config.max\_lag}). \\
   \If(Add connections for noise.){\texttt{graph\_config.include\_noise}}{
    \ForEach{\texttt{noise\_node}}{
      Add edge to its corresponding target, feature, or latent node.
    }
   }
   \If(Add autoregressive edges.){\texttt{graph\_config.max\_lag} > 0}{
    \ForEach{target, feature, latent, and noise variable in graph}{
      \texttt{add\_edge} $\sim$ Bernoulli(\texttt{graph\_config.prob\_<var\_type>\_autoregressive}) \\
      \If{\texttt{add\_edge is True}}{
        Add forward (in time) edge between consecutive nodes of current variable. \\
      }
    }
   }
   \ForEach{Non-noise node at time \texttt{t} (i.e., lag of 0)}{
    Contruct a list of possible parents based on \texttt{graph\_config.min\_lag} (or current topology of DAG if \texttt{min\_lag} is 0 to maintain acyclic graph), \texttt{graph\_config.allow\_latent\_direct\_target\_cause}, and \texttt{graph\_config.allow\_target\_direct\_target\_cause}. \\
    Shuffle list of possible parents. \\
    \While{Current number of parents < \texttt{graph\_config.max\_<var\_type>\_parents} and length of list of possible parents > 0}{
      Pop first element from list of possible parents. \\
      \texttt{prob\_edge} ${\leftarrow}$ \texttt{graph\_config.prob\_<var\_type>\_parent} if not \texttt{None} else \texttt{graph\_config.prob\_edge} \\
      \texttt{add\_edge} $\sim$ Bernoulli(\texttt{prob\_edge}) \quad \# Note: \texttt{prob\_edge} controls graph sparsity. \\
      \If{\texttt{add\_edge is True} and the number of existing children of the current possible parent node < \texttt{graph\_config.max\_<parent\_type>\_children}}{
        Add edge from possible parent to current node. \\
        Add edges between nodes with appropriate lags up to \texttt{graph\_config.max\_lag}, e.g., if $X_j(t-1) \rightarrow X_i(t)$, then $X_j(t-2) \rightarrow X_i(t-1)$, etc. \\
        \If{Number of parents of current node from the same variable $\geq$ \texttt{graph\_config.max\_parents\_per\_variable}}{
          Remove any other instances of the same variable as the current parent from the list of possible parents. For example, this parameters prevents $X_j(t-1) \rightarrow X_i(t)$ and $X_j(t-2) \rightarrow X_i(t)$ if \texttt{graph\_config.max\_parents\_per\_variable} is 1.
        }
      }
    }
   }
   \KwOut{TimeSeriesCausalGraph}
   \caption{Time Series Causal Graph Generation}
   \label{algo:ts_causal_graph_generating_process}
  \end{algorithm}

  \begin{algorithm}[h]
  \SetAlgoLined
  \KwIn{\texttt{function\_config}: FunctionConfig, \texttt{causal\_graph}: TimeSeriesCausalGraph}
   Complete missing values with defaults based on complexity value for \texttt{function\_config}. \\
   \ForEach{Node in \texttt{causal\_graph} at time \texttt{t} (i.e., lag of 0)}{
    \eIf{Current node is a noise node and it has a parent}{
      Sample linear weights for functional dependency $f_i$ between its parent and itself as its only possible parent is the lagged version of itself and this relationship is currently limited to linear. \\
      Set autoregressive relationship to linear function with sampled parameters. \\
    }{
      \ForEach{Parent of current node}{
        \eIf{Current parent is a noise node}{
          Set functional dependency $f_{ij}$ to the identity function as noise is currently simply treated as additive.
        }{
          Randomly sample function type from \texttt{function\_config.functions} with probabilities \texttt{function\_config.prob\_functions}. \\
          Randomly sample parameters for chosen function type (cf. example script for details). \\
          Set functional dependency $f_{ij}$ to sampled function with sampled parameters. \\
        }
      }
    }
   }
   \KwOut{StructuralCausalModel}
   \caption{Structural Causal Model Generation}
   \label{algo:scm_generating_process}
  \end{algorithm}

  \clearpage
  \section{Supplemental experiments} \label{app:supplemental_results}

\begin{figure}[htb]
      \centering
      \subfigure[F1]{
        \centering
        \includegraphics[width=0.46\linewidth]{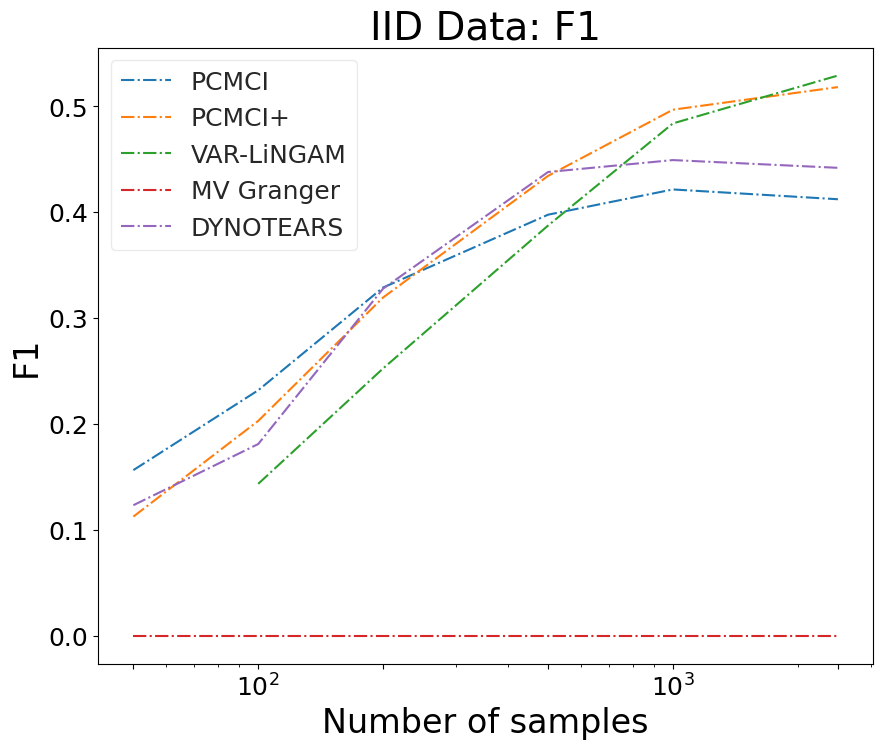}
        \label{fig:f1_iid_data}
      }\hfill
      \subfigure[SHD]{
        \centering
        \includegraphics[width=0.46\linewidth]{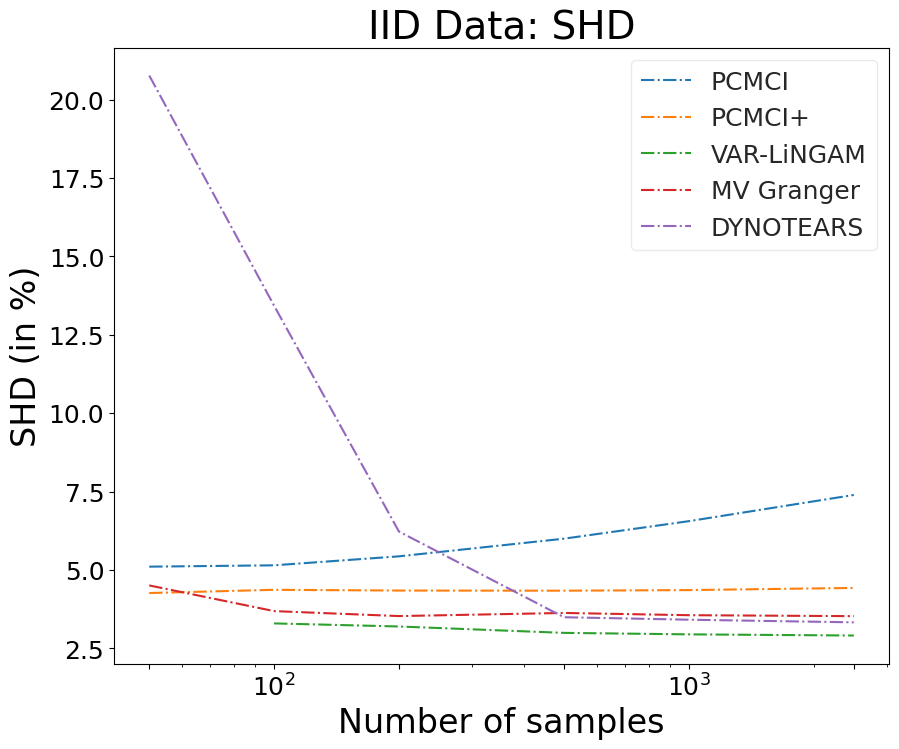}
        \label{fig:shd_iid_data}
      }
      \subfigure[Normalized true positives]{
        \centering
        \includegraphics[width=0.46\linewidth]{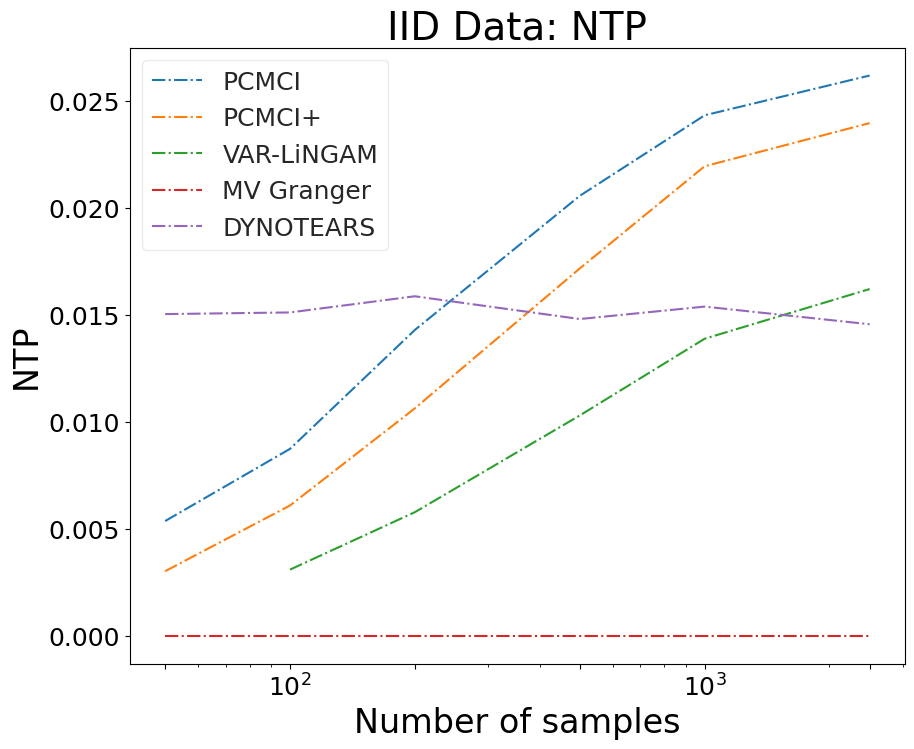}
        \label{fig:ntp_iid_data}
      }\hfill
      \subfigure[Normalized false positives]{
        \centering
        \includegraphics[width=0.46\linewidth]{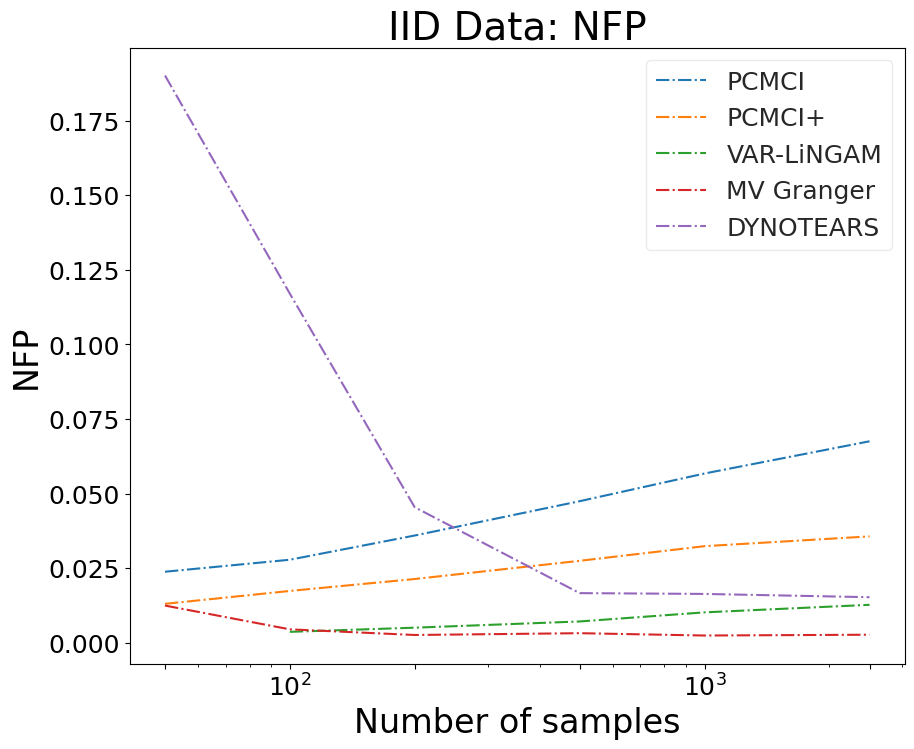}
        \label{fig:nfp_iid_data}
      }
      \caption{IID Data - For IID data the true links in the causal graph only exist between nodes with a relative lag of 0. As described in~\autoref{fig:instantaneous_effects}, our implementation of multivariate Granger can never identify instantaneous effects. For the IID case, all the true edges are instantaneous effects and the method returns zero true positives~\subref{fig:ntp_iid_data}. \subref{fig:f1_iid_data}~F1 improves for all other methods as the number of observations increase. The increase in SHD~\subref{fig:shd_iid_data} for PCMCI is again attributed to the increase in false positives~\subref{fig:nfp_iid_data} due to its inability to orient edges of contemporaneous links as mentioned in~\autoref{fig:instantaneous_effects}.}
      \label{fig:iid_data}
    \end{figure}

\begin{figure}[htb]
      \centering
      \subfigure[F1]{
        \centering
        \includegraphics[width=0.46\linewidth]{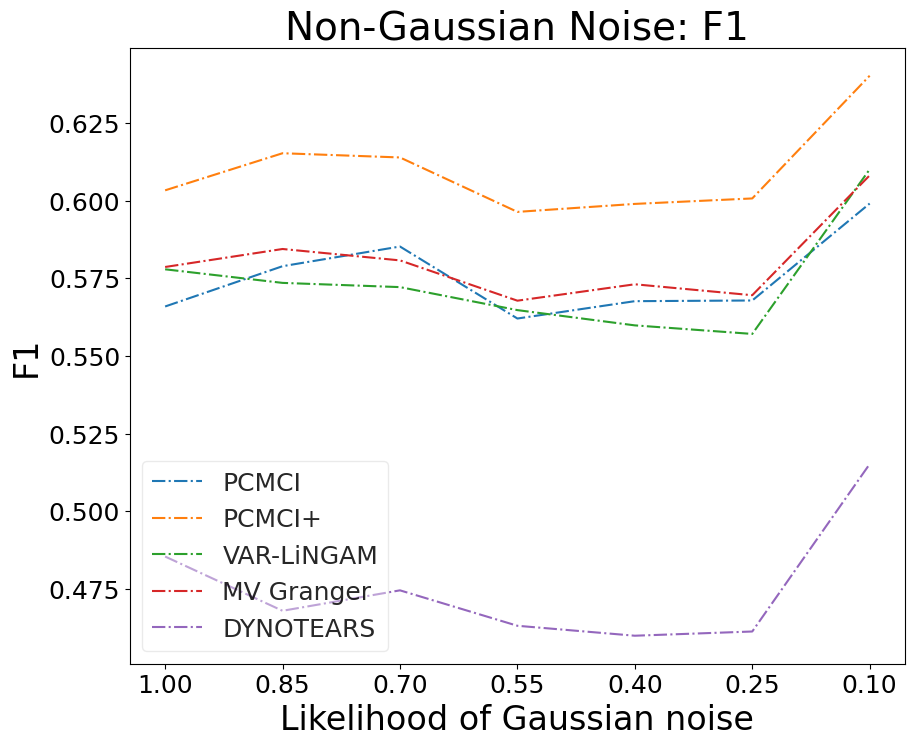}
        \label{fig:f1_nongaussian_noise}
      }\hfill
      \subfigure[SHD]{
        \centering
        \includegraphics[width=0.46\linewidth]{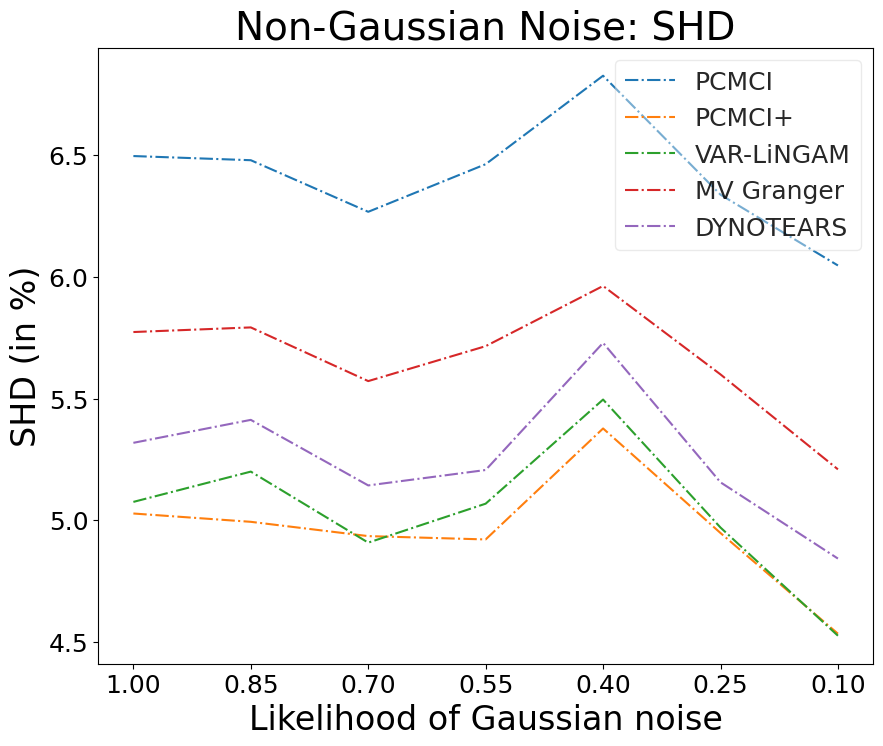}
        \label{fig:shd_nongaussian_noise}
      }
      \subfigure[Normalized true positives]{
        \centering
        \includegraphics[width=0.31\linewidth]{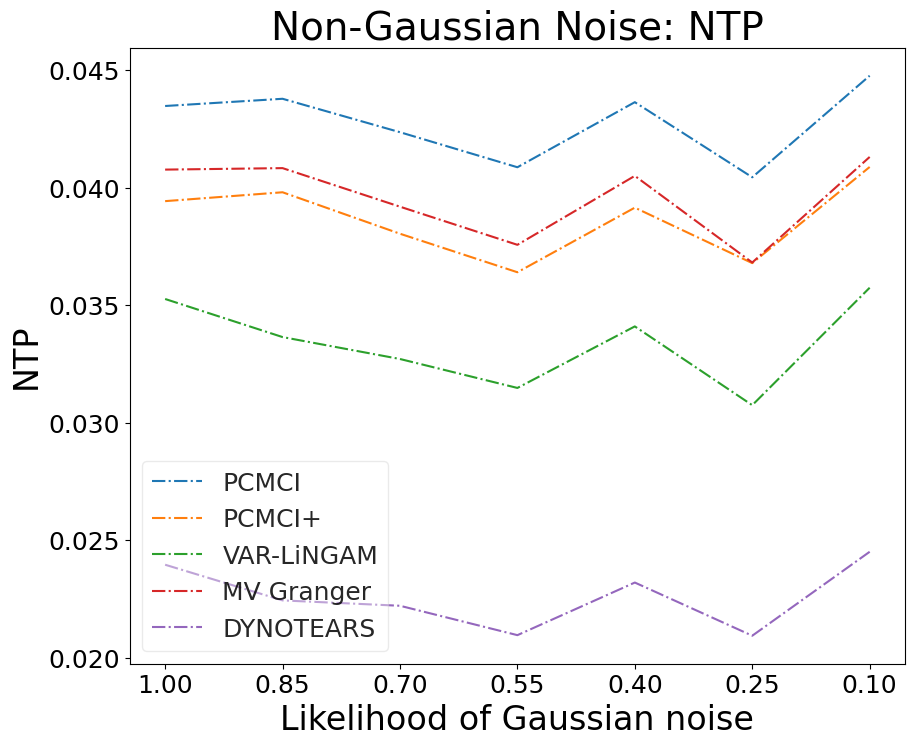}
        \label{fig:ntp_nongaussian_noise}
      }\hfill
      \subfigure[Normalized false positives]{
        \centering
        \includegraphics[width=0.31\linewidth]{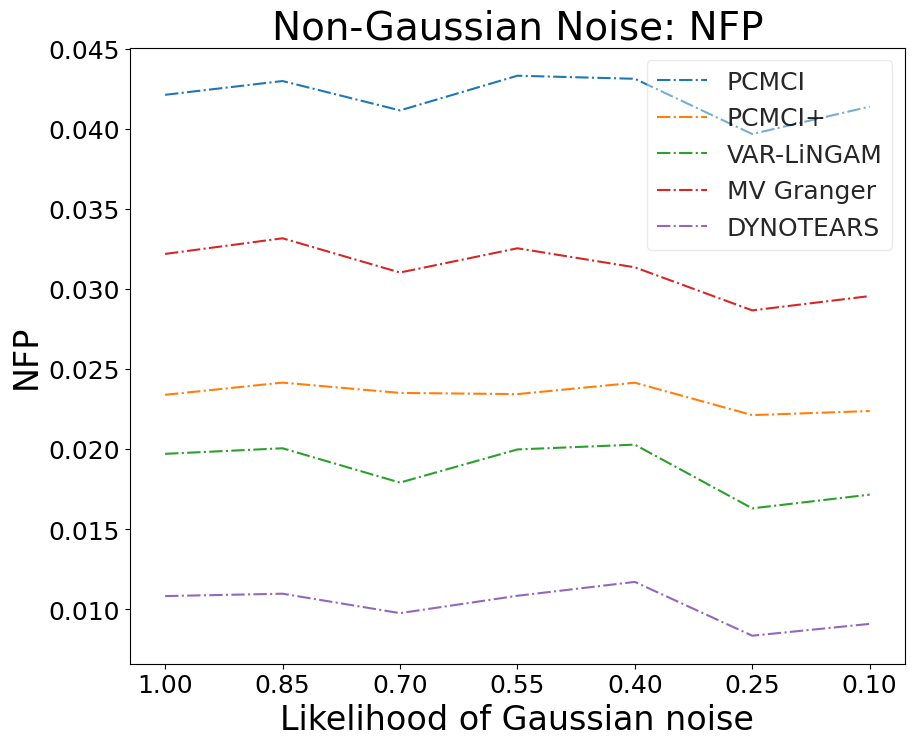}
        \label{fig:nfp_nongaussian_noise}
      }\hfill
      \subfigure[Normalized false negatives]{
        \centering
        \includegraphics[width=0.31\linewidth]{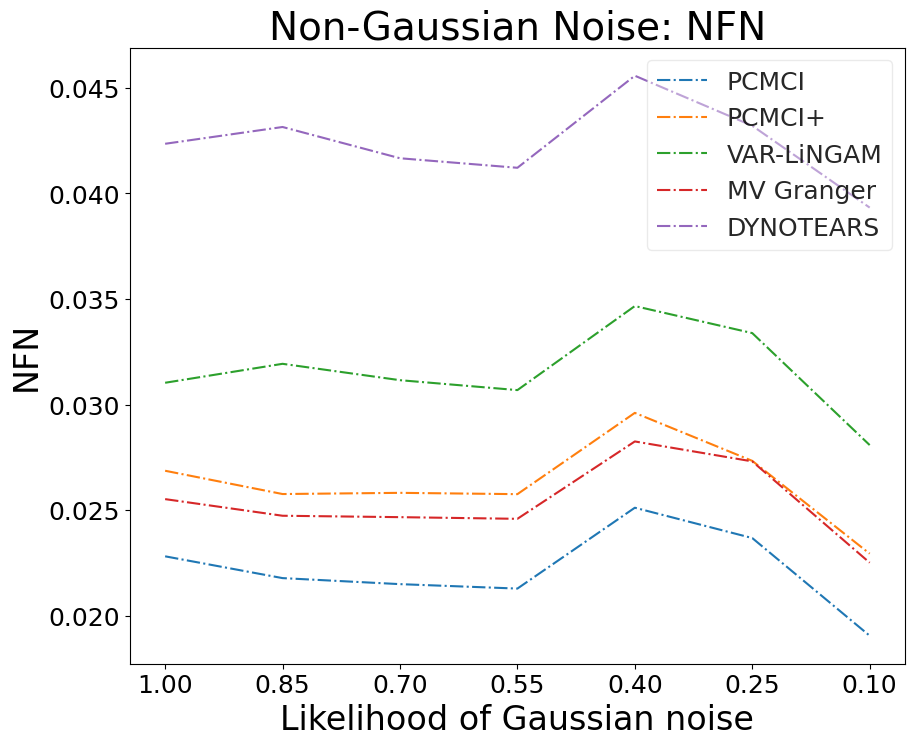}
        \label{fig:nfn_nongaussian_noise}
      }
      \caption{Non-Gaussian Noise - \subref{fig:f1_nongaussian_noise}~F1 is stable for all methods so they perform fairly well when noise is sampled from distributions with fatter tails. The increase in F1 below 25\% can be attributed to both the increase in true positives~\subref{fig:ntp_nongaussian_noise} and the decrease in false negatives~\subref{fig:nfn_nongaussian_noise}. \subref{fig:shd_nongaussian_noise}~SHD follows the trajectory of the false negatives as the false positives~\subref{fig:nfp_nongaussian_noise} are relatively flat in comparison.}
      \label{fig:nongaussian_noise}
    \end{figure}

    We performed two additional experiments, as outlined in~\autoref{table:supplemental_experiments}, on the methods in~\autoref{table:methods}.
    As described in~\autoref{sec:experiments}, for each experiment, 200 unique SCMs were generated from the same parameterization space defined for the specific experiment. For the non-Gaussian noise experiment, a single data set with 1000 samples was generated from each SCM, while we varied the number of samples in the data set from each SCM for the IID experiment.
    The following results shown in~\autoref{fig:iid_data} and~\autoref{fig:nongaussian_noise} capture the average metrics, defined in~\autoref{sec:metrics}, for each causal discovery method across the 200 data sets with known causal ground truth for the experiments defined in~\autoref{table:supplemental_experiments}, respectively.

    \begin{table}[ht]
     \caption{Supplemental experiments}
     \label{table:supplemental_experiments}
     \centering
     \scalebox{0.794}{
       \begin{tabularx}{\textwidth}{lX}
         \toprule
         Name & Description \\
         \midrule
         4. IID Data & Only IID data is generated. \\
         5. Non-Gaussian Noise & The likelihood of additive Gaussian noise is decreased while the likelihood of Laplace, Student's t, and uniform distributed noise is increased. The noise variance remains unchanged. \\
         \bottomrule
       \end{tabularx}
     }
    \end{table}

  \clearpage
  \section{Example configuration} \label{app:example_config}

\begin{lstlisting}[language=Python]
import (CausalGraphConfig, DataGenerationConfig, FunctionConfig, NoiseConfig,
        RuntimeConfig)

# DataGenerationConfig is the top-level configuration object and provides the capability
# to specify all necessary parameters to generate a SCM and sample time series data.
config = DataGenerationConfig(
    # Controls random behavior and ensures reproducibility for graph and SCM generation.
    random_seed=1,
    # Used to initialize any unspecified configurations.
    # They are all initialized in this example so value would be ignored.
    complexity=20,
    percent_missing=0.0,  # Percentage of missing data (NaN values) in final data set(s).
    causal_graph_config=CausalGraphConfig(
        # Used to complete any unspecified parameters of the CausalGraphConfig.
        graph_complexity=20,
        include_noise=True,  # Noise is included in the system. Should always be True.
        max_lag=3,  # Maximum possible lag between a parent and child.
        # Minimum possible lag between a parent and child. 0 allows instantaneous effects
        min_lag=1,
        num_targets=1,  # Number of target variables.
        num_features=10,  # Number of feature variables.
        num_latent=2,  # Number of latent variables.
        # Likelihood of an edge. Used to control graph sparsity.
        # Used when prob_<var_type>_parent is undefined for specific <var_type>.
        prob_edge=0.25,
        # Only 1 lagged node of a variable can be a parent of another node.
        max_parents_per_variable=1,  # Helps to control graph sparsity.
        # max_<var_type>_parents and max_<var_type>_children help control graph sparsity.
        max_target_parents=2,  # Maximum number of parents for a target node.
        max_target_children=0,  # Maximum number of children for a target node.
        # Likelihood of an edge between a possible parent and a target variable.
        prob_target_parent=0.2,  # Helps to control graph sparsity.
        max_feature_parents=3,  # Maximum number of parents for a feature node.
        max_feature_children=2,  # Maximum number of children for a feature node.
        max_latent_parents=2,  # Maximum number of parents for a latent node.
        max_latent_children=2,  # Maximum number of children for a latent node.
        # A latent variable cannot be a direct cause of a target variable.
        allow_latent_direct_target_cause=False,
        # A target variable cannot be a direct cause of another target variable.
        allow_target_direct_target_cause=False,
        # Likelihood of autoregressive relationship for a target variable.
        prob_target_autoregressive=1.0,
        # Likelihood of autoregressive relationship for a feature variable.
        prob_feature_autoregressive=0.8,
        # Likelihood of autoregressive relationship for a latent variable.
        prob_latent_autoregressive=0.5,
        # Likelihood of autoregressive relationship for a noise variable.
        prob_noise_autoregressive=0.1
    ),
    function_config=FunctionConfig(
        # Used to complete any unspecified parameters of the FunctionConfig.
        function_complexity=30,
        # Possible functional dependencies on each edge.
        functions=["linear", "monotonic", "trigonometric"],
        # Likelihood of choosing above functional forms.
        prob_functions=[0.4, 0.5, 0.1]
    ),
    noise_config=NoiseConfig(
        # Used to complete any unspecified parameters of the NoiseConfig.
        noise_complexity=20,
        # Variance of each noise node is drawn from uniform distribution with given range
        noise_variance=[0.01, 0.1],
        # Possible noise distributions.
        distributions=["gaussian", "laplace", "students_t", "uniform"],
        # Likelihood of choosing above distributions.
        prob_distributions=[0.4, 0.2, 0.2, 0.2]
    ),
    runtime_config=RuntimeConfig(
        # Two data sets will be returned. One with 100 and another with 500 observations.
        num_samples=[100, 500],
        # Seeds used to sample from the SCM to produce the two data sets.
        data_generating_seed=[42, 43],
        # Values for latent and noise variables will not be returned.
        return_observed_data_only=True,
        # Data will not be normalized to zero mean and unit variance.
        normalize_data=False
    )
)
\end{lstlisting}

\end{appendices}

\end{document}